\documentclass[authoryear,preprint]{elsarticle}

\journal{Artificial Intelligence in Agriculture}

\usepackage[english]{babel}

\usepackage{amsmath}
\usepackage{amsfonts}
\usepackage{mathrsfs}

\usepackage{graphicx}
\usepackage{booktabs}
\usepackage{caption}
\usepackage{subcaption}
\usepackage{threeparttable}
\usepackage{diagbox}
\usepackage{makecell}
\usepackage{siunitx}

\usepackage{algorithm}
\usepackage{algorithmicx}
\usepackage{algpseudocode}
\usepackage{listings}

\usepackage{enumitem}
\usepackage[title]{appendix}
\usepackage{chngcntr}

\usepackage[T1]{fontenc}
\usepackage{csquotes}
\usepackage{comment}
\usepackage{xcolor}
\usepackage[colorlinks=true, allcolors=blue]{hyperref}
\usepackage{orcidlink}



\begin{document}

\begin{frontmatter}

\title{Field-Level Crop Classification via Similarity-Induced Supervision from Imperfect Labels}

\author[1]{Raul Lomonte}
\ead{raullomonte13@gmail.com}

\author[2]{João Rebouças Dorea\orcidlink{0000-0001-9849-7358}}
\ead{joao.doreaa@wisc.edu}

\author[1]{Dário Augusto Borges Oliveira\orcidlink{0000-0002-0674-5332}}
\ead{dario.oliveira@fgv.br}

\affiliation[1]{organization={School of Applied Mathematics, Getulio Vargas Foundation},
addressline={Praia de Botafogo},
city={Rio de Janeiro},
state={RJ},
country={Brazil}}

\affiliation[2]{organization={Animal Sciences Department, University of Wisconsin-Madison},
city={Madison},
state={WI},
country={USA}}

\begin{abstract}
Regional-scale crop mapping is a critical task for food production monitoring and land use management, yet remains challenging in tropical regions, where persistent cloud cover limits the availability of optical imagery during the growing season. Synthetic aperture radar (SAR), being insensitive to atmospheric conditions, circumvents this problem and enables continuous observations throughout the agricultural cycle. A second obstacle is label quality: land cover products generated at national scale exhibit systematic biases that underestimate less prevalent crops, and training a classifier directly on such labels entails inheriting and amplifying their errors. This work proposes a methodology to build effective large-scale crop classification models directly from imperfect public land cover products, requiring no new field campaign, no additional ground survey, and no manual label correction. The approach combines two complementary mechanisms that act simultaneously on two distinct sources of inconsistency. The first reduces the semantic inconsistency of the labels by shifting supervision from the sample level to the group level: a similarity-based clustering model organizes SAR time series into semantically coherent groups, and noisy labels are used solely to assign each group its most representative class, diluting the impact of individual labeling errors. An analysis of the resulting latent space shows clusters that are highly pure with respect to the reference classes, 89.0\% and 91.2\% in the two study regions, confirming that the representation organizes the crops well independently of the noisy label source. The second reduces the spatial inconsistency of the predictions by exploiting the physical consistency of the agricultural field: a segmentation model automatically delineates crop field boundaries from optical imagery, and patch-level predictions are consolidated by majority voting within each crop field, suppressing intra-field noise and yielding spatially coherent maps. The methodology is evaluated in two Brazilian agricultural municipalities with contrasting characteristics, Campo Verde (Mato Grosso) and Luís Eduardo Magalhães (Bahia), using a national land cover product as the noisy label source and independent field annotations as reference. The proposed approach consistently outperforms the noisy baseline in mean class accuracy, with gains of up to 10.14 percentage points when both mechanisms are combined, particularly pronounced for minority crops. Because every stage relies exclusively on publicly available imagery and labels, the pipeline scales without further annotation: a state-scale generalization analysis, applied to Mato Grosso and Bahia without any further model adjustment, reveals stable behavior across dozens of municipalities. The municipal crop areas derived from these predictions are furthermore consistent with the official Brazilian agricultural statistics (IBGE/PAM), an entirely independent estimate that the model never observes, the two converging onto the identity line in the municipalities where each crop occupies a large area, confirming the viability of the methodology for large-scale crop classification from imperfect labels.
\end{abstract}


\end{frontmatter}



\section{Introduction}

Crop-type classification from remote sensing is an essential component for food production monitoring, land use planning, and environmental impact assessment at regional and national scales. In Brazil, the world's largest soybean producer and one of the leading exporters of cotton and maize, the ability to automatically map crop distribution throughout the agricultural cycle has direct implications for public policies on food security, supply chain traceability, and enforcement against deforestation associated with agricultural expansion. The magnitude of this demand, combined with the diversity of crops, regions, and growing season patterns across the national territory, makes it infeasible to rely exclusively on field-collected annotations: scaling classification with automatically produced labels is necessary.

Initiatives such as MapBiomas address this need by producing, annually, land use and land cover maps for the entire country from satellite image time series combined with machine learning models and prior knowledge. These products offer unprecedented spatial coverage, but carry well-known structural limitations. Being generated at national scale with a 30-meter resolution, their classifications tend to exhibit systematic biases that favor dominant crops, such as soybean, and underestimate less frequent ones, such as maize and cotton. In regions with complex agricultural patterns or a strong second crop, this imprecision is not marginal: for minority crops in certain regions, the number of patches correctly labeled by MapBiomas can be lower by more than an order of magnitude relative to the actual total, as evidenced in this work. Training a supervised model directly on these labels entails inheriting, and potentially amplifying, the errors of the reference source.

A second challenge arises from the structure of patch-level inference itself. Classification using fixed-size image windows treats each landscape fragment in isolation, ignoring the physical structure of agriculture: within a crop field, by definition, a single crop predominates. Pointwise predictions over individual patches are inherently noisy and produce maps with high spatial variability, fragmented and inconsistent with the agronomic logic of the landscape. This problem is compounded in noisy label scenarios, where the training signal is already imprecise at the individual level.

A third obstacle, specific to tropical agricultural regions of Brazil, is persistent cloud cover during the growing season. Optical imagery, such as that produced by the Sentinel-2 satellite, is frequently compromised by clouds in the months of highest agricultural activity, reducing the availability of valid observations precisely when phenological discrimination among crops would be most informative. Synthetic aperture radar (SAR), operating in the microwave band and insensitive to atmospheric conditions, represents a robust alternative: the Sentinel-1 satellite, part of the ESA Copernicus programme, offers systematic coverage and regular revisits regardless of cloud presence, making it especially well-suited for crop monitoring in tropical regions.

Faced with these interrelated challenges, this work proposes a methodology to build effective large-scale crop classification models directly from imperfect public land cover products, reducing simultaneously the semantic inconsistency of the labels and the spatial noise of the predictions. The proposal rests on two complementary mechanisms, each targeting one of these two sources of inconsistency. The first mechanism attacks the semantic inconsistency of the labels through pseudo-group creation: instead of training a classifier directly on the available labels, we first organize the data into semantically coherent clusters in the latent representation space, using a \textit{Masked Siamese Network} (MSN)~\cite{assran2022maskedsiamesenetworkslabelefficient} trained in a fully self-supervised manner on Sentinel-1 SAR image time series. Noisy labels are then used not to supervise individual samples, but to assign each cluster a name based on its most representative class: by attributing the class to the group rather than to the patch, the error associated with individual incorrect labels is diluted, and the systematic bias of the reference source is substantially attenuated. The second mechanism attacks the spatial inconsistency of the predictions through crop field aggregation: we use the \textit{Segment Anything 2} (SAM2)~\cite{ravi2024sam2segmentimages} model to automatically delineate crop field boundaries from Sentinel-2 median composites, and consolidate individual patch predictions within each crop field by majority voting. By exploiting the physical consistency of the agricultural field unit, in which a single crop predominates, this step suppresses intra-field noise and produces classifications that are spatially coherent with the landscape and robust to pointwise errors. Both mechanisms are evaluated independently and in combination, acting on distinct and complementary sources of inconsistency. Crucially, the entire pipeline operates on publicly available imagery and labels alone: it requires no new field campaign, no additional ground survey, and no manual label correction, which makes it inherently scalable to municipal and state extents.

The methodology is validated on two datasets from Brazilian agricultural regions with contrasting characteristics. The first is the municipality of Campo Verde (CV), in Mato Grosso, a reference region for soybean, cotton, and maize cultivation in the Cerrado, with 513 reference polygons covering 628.72~km². The second is the municipality of Luís Eduardo Magalhães (LEM), in western Bahia, with 1,069 reference polygons and 1,276.59~km² labeled, where second-crop diversity and bi-seasonal patterns introduce additional challenges. In both regions, MapBiomas predictions are used as the noisy training label source, and evaluation is conducted against independent reference annotations. To assess the generalization capability of the methodology beyond the labeled areas, we apply the full pipeline to the entire state of Mato Grosso and to the agricultural region of Bahia, without any further model adjustment. At this scale, the resulting municipal crop areas are further contrasted with the official Brazilian agricultural statistics published by IBGE, a source built from local declarations rather than from imagery and therefore fully independent of every input used by the pipeline.

Results demonstrate that the proposed approach consistently outperforms MapBiomas in mean class accuracy: the pseudo-group mechanism alone yields gains of up to 9.86 percentage points in LEM at the patch level, and combining both mechanisms raises mean class accuracy by 9.91 percentage points in CV and 10.14 in LEM relative to the MapBiomas baseline. Gains are especially pronounced for minority crops, precisely where the bias of the noisy baseline is most severe. Automatic crop field segmentation by SAM2 expands the mapped area by approximately six times relative to the reference annotation coverage in CV, and 2.6 times in LEM, enabling application at the municipal and state scales. The generalization analysis reveals stable behavior across municipalities, with a consistent discordance distribution even considering the heterogeneity of the evaluated regions, and the municipal crop areas estimated at this scale are correlated with the official IBGE statistics, the two estimates converging onto the identity line in the municipalities where each crop occupies a large area.

The contributions of this work are as follows. First, we propose and validate self-supervised pseudo-groups as a mechanism to reduce the semantic inconsistency of land cover labels, demonstrating that the latent organization of the data can replace direct supervision and outperform a biased label baseline. We further characterize the latent space that sustains this mechanism, showing that its clusters are highly pure with respect to the true classes, which is what makes group-level naming sound. Second, we integrate automatic crop field segmentation from optical imagery into the SAR-based classification pipeline as a mechanism to reduce the spatial inconsistency of the predictions, suppressing intra-field noise through majority voting and producing spatially coherent maps, while enabling inference at scale without any additional reference polygon. Third, we show that both mechanisms are complementary, acting on distinct sources of inconsistency, and evaluate their combination in real-world scenarios with different crops, agricultural calendars, and label qualities. Fourth, we demonstrate that the methodology is inherently scalable: because it relies solely on public imagery and public labels and requires no new field campaign, ground survey, or manual label correction, it can be applied to entire states with no further model adjustment, exhibiting controlled discordance variance across dozens of municipalities and therefore robust and predictable behavior beyond the validation regions. Finally, we contrast the resulting maps with the official agricultural statistics of Brazil, an independent estimate built from local declarations rather than from imagery, showing that the municipal crop areas derived from the predictions are consistent with the official figures wherever a crop is a substantial part of the local agricultural landscape.

The remainder of this article is organized as follows. Section~\ref{sec:related} reviews related work on crop classification with noisy labels, self-supervised representation learning, and crop field segmentation. Section~\ref{sec:methodology} describes the proposed methodology. Section~\ref{sec:study_areas} presents the study areas and datasets used. Section~\ref{sec:experiments} details the experiments and discusses the results. Section~\ref{sec:conclusion} presents the conclusions and future perspectives.

\section{Related Works}
\label{sec:related}

The proposed methodology lies at the intersection of four research lines: crop classification from synthetic aperture radar (SAR) time series, self-supervised representation learning for Earth observation, automatic crop field delineation by foundation segmentation models, and training classifiers under noisy labels derived from national-scale land cover products. This section organizes the literature along these axes and highlights the gaps that motivate our contributions.

\subsection{Crop Classification with SAR in Tropical Regions}
\label{subsec:rw_sar}

Agricultural monitoring by remote sensing in tropical regions is strongly limited by cloud cover during the growing season, which compromises the availability of optical observations during the most discriminative phenological windows. Being insensitive to atmospheric conditions, SAR has established itself as a robust alternative in this scenario, and the Sentinel-1 programme has made systematic access to dense dual-polarization image series viable. In the Brazilian context, the body of work developed around the Campo Verde and Luís Eduardo Magalhães datasets established benchmarks directly related to ours. The \emph{Campo Verde} dataset~\cite{sanches2018}, with monthly labeled crop fields over the Cerrado of Mato Grosso, was designed to advance agricultural remote sensing in tropical areas with first-crop and second-crop systems. On this domain, \cite{cuelarosa2019} compared different convolutional architectures on Sentinel-1 multitemporal series, incorporating prior knowledge of the agricultural calendar, while \cite{chamorro2021} achieved state-of-the-art results with fully convolutional recurrent networks evaluated on Campo Verde and Luís Eduardo Magalhães. Recent work continues to demonstrate that temporal architectures operating exclusively on Sentinel-1 discriminate multiple crops with competitive accuracy~\cite{tsuchiya2025}, reinforcing the viability of SAR as the primary input. These studies assume, however, the availability of reliable field labels for direct supervision, an assumption our approach does not require.

\subsection{Self-Supervised Learning for Earth Observation}
\label{subsec:rw_ssl}

Self-supervised representation learning has become the dominant paradigm for reducing dependence on labeled data. Joint-embedding approaches such as SimCLR~\cite{chen2020simclr}, MoCo~\cite{he2020moco}, and DINO~\cite{caron2021dino} demonstrated that representations competitive with supervised learning can be obtained without labels, while masked autoencoders (MAE)~\cite{he2022mae} proposed learning representations by reconstructing masked regions, without relying on contrastive objectives. \emph{Masked Siamese Networks} (MSN)~\cite{assran2022maskedsiamesenetworkslabelefficient}, adopted in this work, occupy a hybrid position between these paradigms, combining patch masking with a siamese prototype assignment structure and achieving particularly strong performance in few-label regimes, a characteristic well aligned with our scenario of scarce and noisy supervision. The transposition of these methods to Earth observation gave rise to a family of self-supervised pre-training approaches for remote sensing, such as SeCo~\cite{manas2021seco}, SatMAE~\cite{cong2022satmae}, and Scale-MAE~\cite{reed2023scalemae}, supported by large-scale multimodal datasets such as SSL4EO-S12~\cite{wang2023ssl4eo} and by pixel-wise time series models such as Presto~\cite{tseng2023presto}. Despite this proliferation, the direct application of MSN to SAR time series, and the use of the resulting cluster assignments as an explicit label noise mitigation mechanism rather than mere initialization for supervised fine-tuning, remains underexplored.

\subsection{Crop Field Delineation and Foundation Segmentation Models}
\label{subsec:rw_sam}

Patch-level inference produces spatially fragmented maps that are inconsistent with the agronomic logic of the landscape, in which the crop field constitutes the natural physical unit of cultivation. Consolidating predictions at this level depends on the availability of crop field geometries, traditionally obtained by manual annotation or by dedicated segmentation networks. The \emph{Segment Anything Model} (SAM)~\cite{kirillov2023sam} and its successor SAM2~\cite{ravi2024sam2segmentimages} changed this landscape by offering zero-shot instance segmentation without domain-specific labels, with SAM2 being more efficient and accurate thanks to a unified memory-capable architecture. Several studies have investigated their adaptation to crop field delineation: \cite{tripathy2024} evaluated SAM in zero-shot mode for smallholder farms, \cite{ha2026} employed seasonal Sentinel-2 composites as input, and methods such as fabSAM~\cite{xie2025fabsam} and FieldSeg~\cite{sun2025fieldseg} achieved additional gains with learned prompters or decoder fine-tuning. These studies focus on segmentation quality as an end in itself. Our work differs by employing SAM2 as a means, using the segmented geometries as support for majority-voting prediction aggregation and integrating optical segmentation into a SAR-based classification pipeline.

\subsection{Large-Scale Mapping and Learning under Noisy Labels}
\label{subsec:rw_noisy}

National-scale land use and land cover mapping initiatives, such as MapBiomas~\cite{MapBiomas/2XSLB3_2025}, offer unprecedented spatial coverage but carry well-documented structural limitations. Their 30-meter resolution and national scale of operation introduce systematic biases that favor dominant crops and underestimate less frequent ones, as well as temporal inconsistencies such as spurious class alternation across years~\cite{rosa2025alternation}. Training a supervised model directly on these labels entails inheriting, and potentially amplifying, the errors of the reference source. This problem has motivated a growing body of work on learning from noisy or weak labels in remote sensing. One family of approaches acts directly on the labels, detecting and correcting dubious pixel-level annotations, for example by clustering pixels in the feature space to estimate per-class prototypes and then reassigning or discarding inconsistent labels~\cite{fritsch2024}. Others keep the labels intact but make training robust to noise, either by exploiting model consensus~\cite{aksoy2022consensual} or by resorting to fault-tolerant learning to transfer coarse labels to finer-resolution predictions~\cite{han2022faulttolerant}. Closest to the object-based half of our proposal, \cite{krishnamurthy2024samodified} employ zero-shot segmentation to delineate parcels and reassign uncertain pixels based on the local label statistics of each segment. Closest to our group-level supervision strategy, and also evaluated on two Brazilian agricultural regions, \cite{larosa2022proportions} use regional proportions of government labels as priors for an online deep clustering method applied to SAR and optical imagery, demonstrating that weak aggregate-level priors can replace per-sample annotation. Our contribution distinguishes itself by combining these two directions, which in isolation attack distinct sources of inconsistency: rather than correcting labels pixel by pixel or training a robust classifier on them, we shift the supervision unit itself from the sample level to the group level to reduce semantic inconsistency, using MapBiomas solely to name self-supervised clusters, and then reduce the residual spatial inconsistency through crop field aggregation. Critically, both directions are realized using only public satellite imagery and public labels, so the resulting pipeline requires no new field campaign, ground survey, or manual relabeling and remains applicable at scale. Evaluating this combination in contrasting regions, as well as applying it at the state scale without further adjustment, fills an underexplored gap in the literature: the integration of self-supervised learning on SAR with object-based aggregation for large-scale crop classification under imperfect labels.

\section{Methodology}
\label{sec:methodology}

Building crop classification models at large scale from public land cover products faces two central and interrelated forms of inconsistency. The first is semantic inconsistency in the labels: land cover databases produced at national scale exhibit systematic classification errors, particularly for less prevalent crops. The second is spatial inconsistency in the predictions: pointwise inferences over small image windows tend to be noisy and fragmented when analyzed individually, conflicting with the agronomic logic of the landscape.

\begin{figure}[ht]
    \centering
    \includegraphics[width=\linewidth]{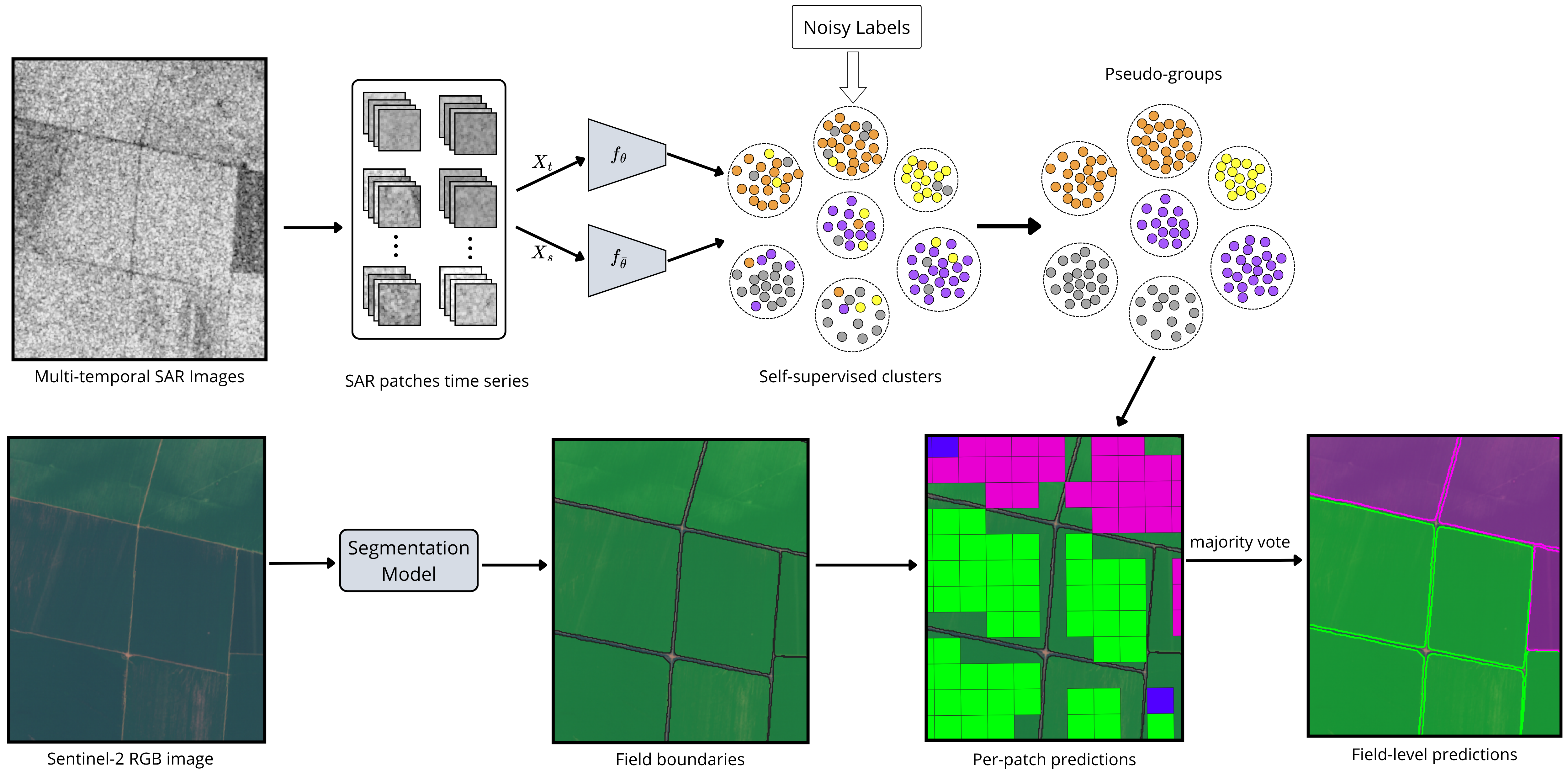}
    \caption{Overview of the proposed methodology for crop-type classification under label noise. \textbf{(Top)} Pseudo-group creation pipeline: a multi-temporal series of SAR patches from the target region is fed into a similarity-based clustering model, where the views $X_t$ and $X_s$ are processed by the encoders $f_\theta$ and $f_{\bar{\theta}}$ to produce self-supervised cluster assignments. Noisy labels are then used to assign a crop class to each cluster based on its most representative label, yielding pseudo-groups that mitigate the impact of label noise. \textbf{(Bottom)} Crop field aggregation pipeline: a Sentinel-2 RGB image of the same region is passed through a segmentation model to delineate crop field boundaries. Per-patch predictions from the upstream classifier are then aggregated within each crop field by majority voting, producing spatially consistent crop-type maps. Colors in the final map denote different crop classes.}
    \label{fig:methodology_overview}
\end{figure}

To mitigate both effects in a scalable manner, and without resorting to new field campaigns or manual label correction, we propose two complementary mechanisms, illustrated in Figure~\ref{fig:methodology_overview}: (i) pseudo-group creation, which targets the semantic inconsistency of the labels by organizing data into semantically coherent clusters from self-supervised representations and naming groups rather than individual samples; and (ii) crop field aggregation, which targets the spatial inconsistency of the predictions by exploiting the physical consistency of the agricultural field unit to consolidate multiple noisy patch predictions into a single, spatially coherent classification. Both mechanisms operate exclusively on publicly available imagery and labels, which is what allows the pipeline to scale to municipal and state extents.

Figure~\ref{fig:methodology_overview} synthesizes the proposed methodological pipeline. In the upper pipeline, SAR image time series are cropped into patches and fed into a self-supervised clustering model (MSN), whose encoders $f_\theta$ and $f_{\bar{\theta}}$ learn cluster assignments from complementary views $X_t$ and $X_s$ of the same input. Noisy labels are then used to assign each cluster its most representative class, forming the pseudo-groups. In the lower pipeline, an RGB image from the Sentinel-2 satellite is passed through a segmentation model (SAM2) to automatically delineate crop field boundaries. The predictions obtained via pseudo-groups are then aggregated within each crop field by majority voting, producing spatially consistent crop maps.

Each approach is evaluated independently, and we subsequently investigate the effect of their combination.

\subsection{Pseudo-Group Creation}

Label noise is a structural limitation of any land cover database produced at scale. Automatically generated labels tend to exhibit systematic biases, overclassifying dominant crops and underestimating less frequent ones. Training a supervised model directly on these labels entails inheriting and potentially amplifying these errors.

Our approach circumvents this problem by decoupling the data organization process from the classification process. Rather than fitting a classifier directly on the available labels, we first organize the data into semantically coherent latent groups, the so-called pseudo-groups, where each group ideally corresponds to a single crop type in the representation space. Only then are noisy labels used, but now to name groups rather than individual samples: by assigning each group its most representative class, the error from individual incorrect labels is diluted, and its impact on the final classification is substantially reduced. The mechanism therefore depends on a premise, namely that the representation is coherent enough for a group to admit a single name; Section~\ref{subsubsec:latent} verifies this premise directly on both study regions.

To build these representations, we use imagery from the Sentinel-1 satellite (ESA Copernicus programme), which operates in the C-band of synthetic aperture radar (SAR). The choice of SAR imagery is justified by its robustness to cloud cover, a frequent phenomenon in tropical agricultural regions during the growing season, when optical imagery is often unavailable. The model input consists of a time series of SAR images concatenated along the agricultural cycle, embedding the phenological dynamics of crops into the representation of each patch.

To learn robust representations, we use a \textit{Masked Siamese Network} (MSN)~\cite{assran2022maskedsiamesenetworkslabelefficient}, a self-supervised clustering model that produces coherent cluster assignments without any label dependency during training. The encoders $f_\theta$ and $f_{\bar{\theta}}$ process pairs of views of the same input, inducing a stable grouping structure in the latent space.

\subsection{Crop Field Aggregation}

The second mechanism targets the spatial inconsistency of patch-level inference, exploiting a fundamental physical property of agriculture: within a crop field, only one crop is predominantly cultivated. Patch-level predictions, taken in isolation, ignore this property and yield fragmented maps in which a single field may receive several conflicting labels. By treating the crop field as the natural unit of decision, this internal consistency can be leveraged to suppress intra-field noise and consolidate pointwise, inherently noisy predictions into a single, spatially coherent classification for the field as a whole.

To apply this strategy at scale, crop field geometries must be obtained automatically from satellite imagery. To this end, we use the \textit{Segment Anything 2} (SAM2)~\cite{ravi2024sam2segmentimages} model, a zero-shot instance segmentation model applied to Sentinel-2 median composites (10-meter spatial resolution), built over the growing season to minimize the influence of clouds and seasonal variations. This automatic segmentation step enables the mapped area to be expanded beyond the available labeled polygons, making classification at the municipal and state scales viable.

After mask generation, a post-processing step is applied to remove regions with area below a minimum threshold and to resolve overlaps between adjacent polygons arising from the division of the image into patches. With the geometries defined, individual patch predictions are consolidated within each crop field by majority voting: the most frequent crop among the inferences associated with the field is assigned to the field as a whole. This spatial smoothing mechanism reduces the impact of pointwise errors and promotes more stable and physically coherent classifications.

\subsection{Combining Both Approaches}

Finally, we evaluate the effect of combining both mechanisms. In this scenario, pseudo-groups produce predictions that are more robust to label noise at the patch level, while crop field aggregation eliminates residual spatial inconsistencies. The two mechanisms act in a complementary manner on distinct sources of inconsistency: the first reduces semantic inconsistency in the labels; the second reduces spatial inconsistency in the predictions. Their combination yields more stable and generalizable classifications, even when facing reference baselines of variable quality. Because both rely solely on public satellite imagery and public labels, the combined pipeline preserves the central property that makes it attractive in practice: it can be deployed over new regions and years without any new field campaign, ground survey, or manual relabeling.

The described methodology was conceived for crop-type classification but is adaptable to other domains where the availability of reliable labels is limited, such as forest monitoring, land use mapping, and vegetation cover analysis.

\section{Study Areas}
\label{sec:study_areas}

The experiments were conducted on two publicly available datasets from Brazilian agricultural regions.

\begin{figure}[H]
    \centering
    \begin{subfigure}{0.48\textwidth}
        \centering
        \includegraphics[width=\linewidth]{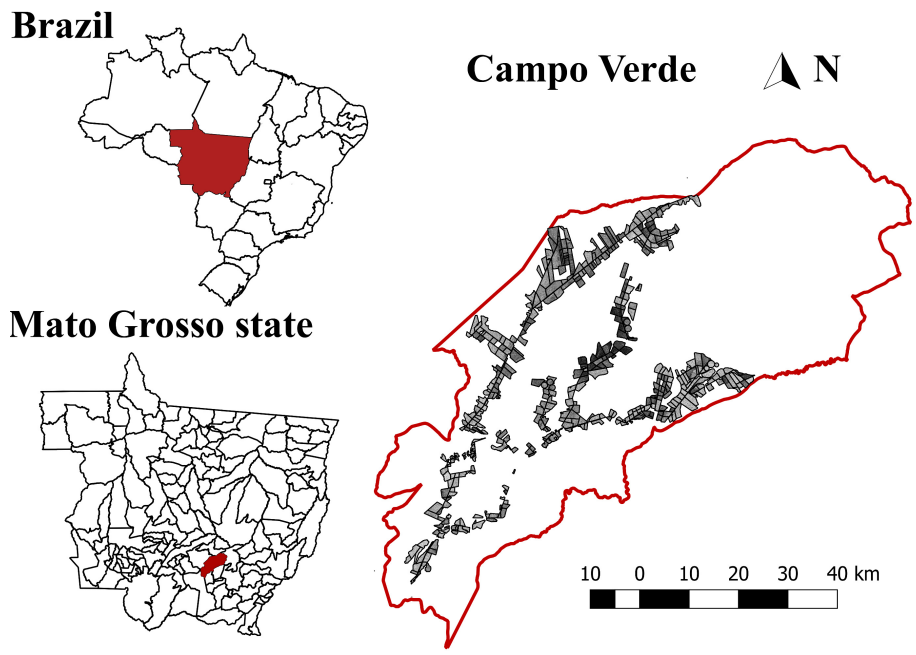}
        \caption{Location of images in CV.}
    \end{subfigure}
    \hfill
    \begin{subfigure}{0.48\textwidth}
        \centering
        \includegraphics[width=\linewidth]{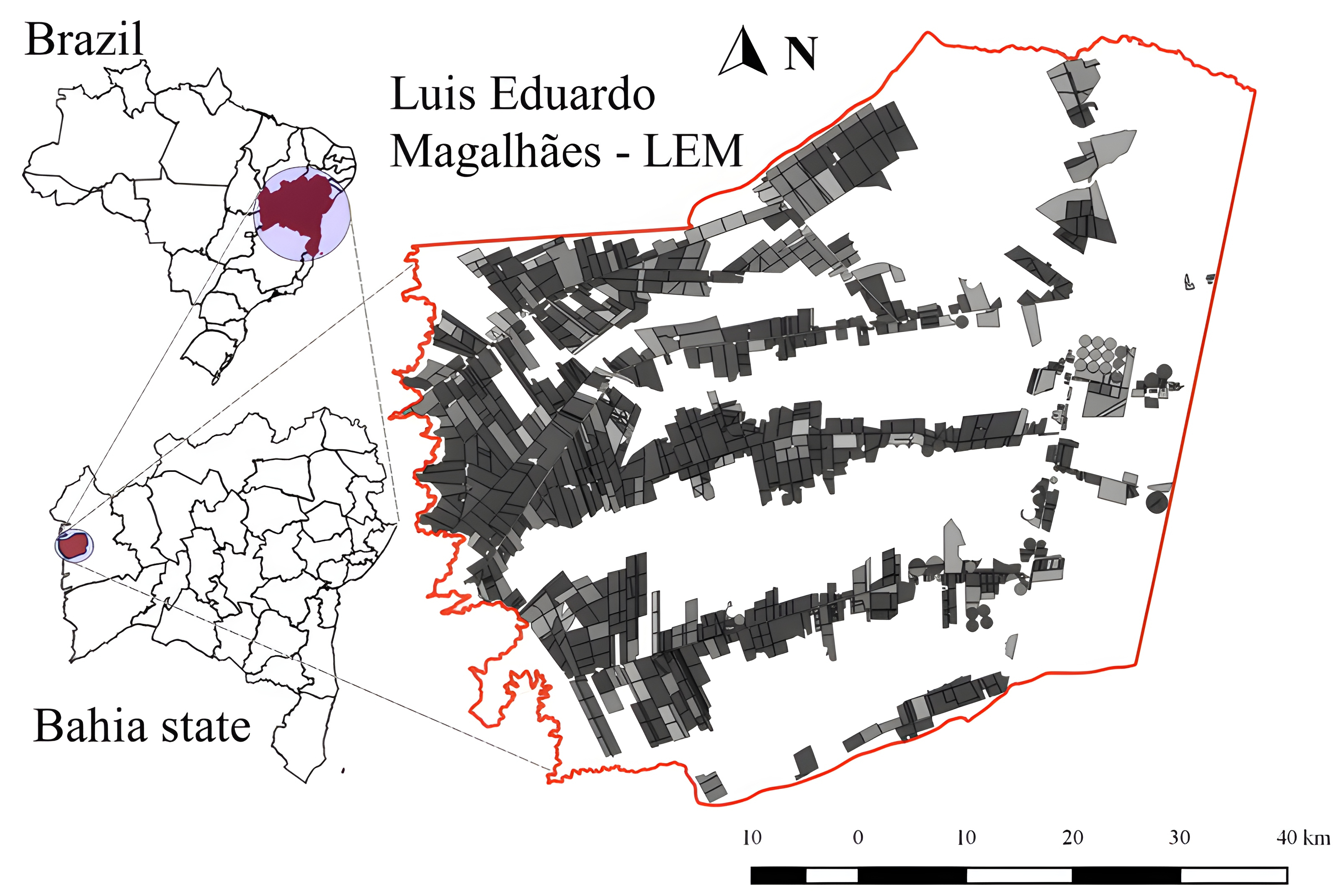}
        \caption{Location of images in LEM.}
    \end{subfigure}
    \caption{Location of the study areas and their reference crop field polygons.}
    \label{fig:loc_dataset}
\end{figure}

\begin{figure}[H]
        \centering
        \includegraphics[width=\linewidth]{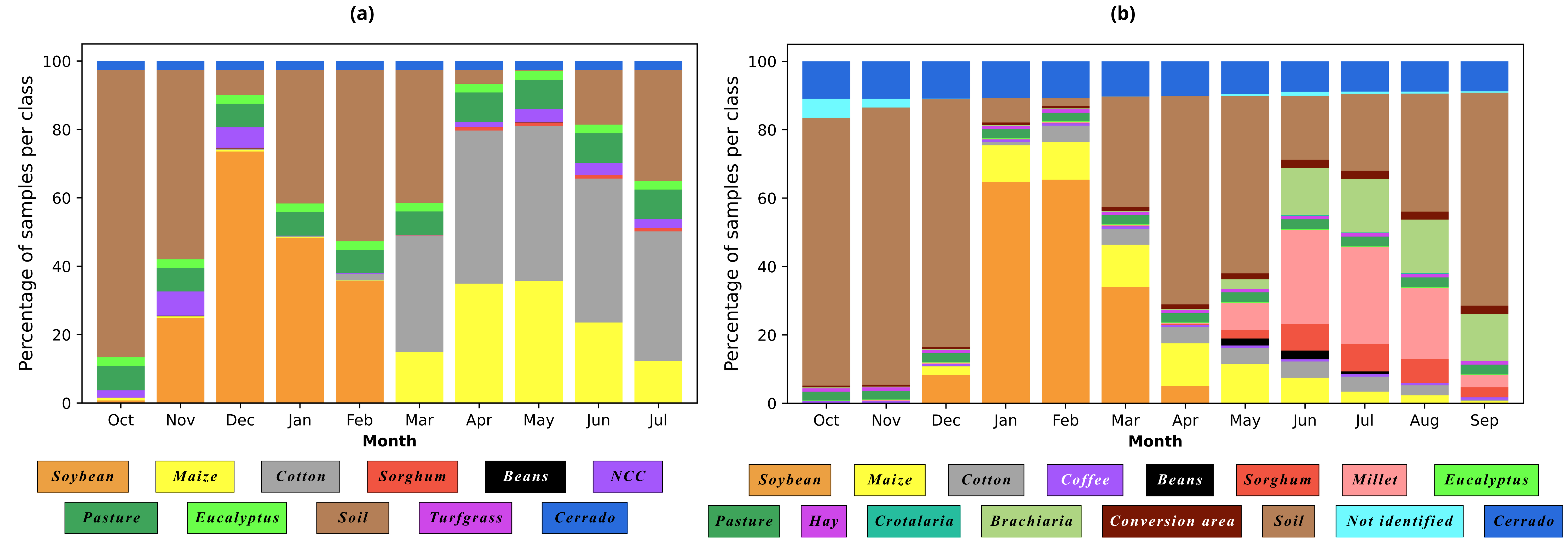}
    \caption{Class distributions for the CV (left) and LEM (right) datasets. NCC refers to Non-Commercial Crops.}
    \label{fig:prop_dataset}
\end{figure}

The first study area is the municipality of Campo Verde (CV), in the state of Mato Grosso, located at latitude 15°32'48"S and longitude 55°10'08"W, as shown in Figure~\ref{fig:loc_dataset}. The municipality covers an area of 4,782~km² [\cite{delarco2018}]. The dataset contains 513 reference polygons, corresponding to 628.72~km² of segmented and classified area. The labels cover the period from October 2015 to July 2016, so the class distribution varies considerably over time (Figure~\ref{fig:prop_dataset}). Soybean was the primary crop cultivated between October 2015 and February 2016, giving way to cotton and maize in the following months.

The second study area is the municipality of Luís Eduardo Magalhães (LEM), in the state of Bahia. This region covers a total area of approximately 3,940~km², centered at coordinates 12°05'31"S and 45°48'18"W. The reference dataset also covers neighboring municipalities, although in this work the analysis is restricted to the municipality of LEM (Figure~\ref{fig:loc_dataset}) [\cite{chamorro2021}]. Restricted to the municipality of LEM, the dataset contains 1,069 reference polygons, representing 1,276.59~km² of segmented and labeled area, covering the period from October 2019 to September 2020. As observed in CV, the temporal class distribution in LEM presents significant variations throughout the year, as illustrated in Figure~\ref{fig:prop_dataset}. The main crops identified in this region were soybean, cultivated between December 2019 and April 2020, and, during the second crop, millet, maize, cotton, and sorghum, among other species. During the second crop, however, only a small fraction of the area was used for cultivation, with much of it remaining as bare soil or brachiaria cover.

\section{Experiments and Discussion}
\label{sec:experiments}

\subsection{Pseudo-Groups}

\subsubsection{Data}

The time series used in this step were obtained from the Sentinel-1 collection available on the Google Earth Engine (GEE) platform. Each image contains two polarization bands, VH and VV. To build the model input, we concatenated five temporally ordered images, generating a 10-band series that captures the spectral dynamics of the agricultural cycle over a five-month period.

The temporal windows were defined according to the agricultural calendar of each region. In CV, the first crop was represented by images from October 2015 to February 2016, aligned with the December 2015 reference labels; the second crop used images from March to July 2016, corresponding to the May 2016 reference labels. In LEM, the first crop comprised images from December 2019 to April 2020, with a February 2020 reference; the second crop was covered by images from May to September, with a July reference period.

As the noisy label source, we used MapBiomas predictions~\cite{MapBiomas/2XSLB3_2025}. MapBiomas is a collaborative initiative bringing together non-governmental organizations, universities, and technology companies, dedicated to the annual production of land use and land cover maps for Brazil. Classifications are generated from satellite image time series combined with machine learning models and prior knowledge, resulting in national-scale products with a spatial resolution of 30~×~30 meters. Although MapBiomas includes agricultural classes such as soybean, cotton, and maize, its predictions present well-known limitations for less prevalent crops, making it an inherently noisy label source.

We used the 2016 data for CV and the 2020 data for LEM (version 10.1), aggregating the land cover classes into four categories: Soybean, Cotton, Maize, and Others, the latter comprising all remaining classes. Table~\ref{tab:patches_mb_inpe} shows the patch count per class in each labeled region, revealing substantial discrepancies between MapBiomas predictions and the annotated ground truth labels, particularly for maize and cotton in LEM.

\begin{table}[h!]
\centering
\small
\caption{Patch count per class for MapBiomas (MB) and ground truth labels in CV and LEM.}
\begin{tabular}{lcccc}
\hline
\textbf{Crop Type}
& \makecell{\textbf{CV} \\ \textbf{MapBiomas}}
& \makecell{\textbf{CV} \\ \textbf{Ground Truth}}
& \makecell{\textbf{LEM} \\ \textbf{MapBiomas}}
& \makecell{\textbf{LEM} \\ \textbf{Ground Truth}} \\
\hline
Soybean    & 5058 & 4359  & 9200  & 8052  \\
Maize   & 2211 & 2131  & 64    & 1836  \\
Cotton & 1620 & 2684  & 669   & 1133  \\
Others  & 2878 & 2593  & 14761 & 13673 \\
\hline
\end{tabular}
\label{tab:patches_mb_inpe}
\end{table}

\subsubsection{Experimental Design}

For the clustering step, we used a Vision Transformer (ViT-S/4) based on the DeiT architecture~\cite{touvron2021trainingdataefficientimagetransformers}, adapted to accept 10-channel time series with a patch size of 4 pixels. MSN projects the representations into a 256-dimensional latent space.

The model was trained with the AdamW optimizer, with an initial learning rate of \num{2e-5} decaying to \num{1e-7} after a 15-epoch warm-up, and weight decay of 0.04 with a final value of 0.4, over 200 epochs. Input patches have a spatial size of 30×30 pixels, with 15\% patch drop. The views used by MSN consist of two 16×16 random views and eight 12×12 focal views. The applied augmentations were random rotation, horizontal flipping, and Gaussian blur; geometric distortion and color manipulation transforms present in the original MSN configuration were removed as they are not appropriate for SAR imagery.

The number of prototypes is the main hyperparameter of this step. To analyze its influence on the results, we trained models with 16, 32, 64, 128, 256, 512, and 1,024 prototypes.

After training, each prototype is classified using MapBiomas labels as reference. To handle class imbalance, rather than assigning each prototype the class with the highest absolute sample count in its neighborhood, we classify each prototype by the class with the highest percentage of samples. This strategy gives greater relative weight to minority classes, which would otherwise be systematically suppressed by the sample volume of dominant classes, a problem especially pronounced in LEM.

Performance is evaluated by overall accuracy (OA) and mean class accuracy (MA), defined as:

\begin{equation}
    \text{OA} = \frac{\sum_{c} n_{cc}}{\sum_{c} \sum_{c'} n_{cc'}}
\end{equation}

\begin{equation}
    \text{MA} = \frac{1}{C} \sum_{c} \frac{n_{cc}}{\sum_{c'} n_{cc'}}
\end{equation}

\noindent where $n_{cc}$ is the number of correctly classified samples for class $c$, $n_{cc'}$ is the number of class-$c$ samples predicted as $c'$, and $C$ is the total number of classes. Mean class accuracy is particularly relevant in this context, as it weights all classes equally regardless of their sample volume.

To characterize the latent space on which the pseudo-groups are built, we additionally report two descriptors computed over the MSN embeddings. Both are measured against the reference annotations rather than against MapBiomas, since the purpose is to assess how well the representation organizes the true classes. The first descriptor is the mean pairwise cosine distance between the embeddings of two classes,

\begin{equation}
    d(c, c') = \frac{1}{|S_c|\,|S_{c'}|} \sum_{i \in S_c} \sum_{j \in S_{c'}} \left( 1 - \frac{z_i \cdot z_j}{\lVert z_i \rVert \, \lVert z_j \rVert} \right),
\end{equation}

\noindent where $S_c$ is the set of patches of class $c$ and $z_i$ is the embedding of patch $i$. The diagonal $d(c,c)$ measures the intra-class dispersion of a crop, and the off-diagonal terms measure its separation from the remaining crops. The second descriptor is cluster purity, the fraction of patches in a cluster that belong to the majority reference class of that cluster,

\begin{equation}
    \text{purity}(k) = \frac{1}{|S_k|} \max_{c} \left| \{\, i \in S_k : y_i = c \,\} \right|,
\end{equation}

\noindent where $S_k$ is the set of patches assigned to cluster $k$ and $y_i$ is the reference label of patch $i$. Global purity is the patch-weighted mean of $\text{purity}(k)$ over all clusters and summarizes, in a single number, how consistently the clusters align with the true classes. This characterization is diagnostic and is only possible where reference annotations exist: it plays no part in the pipeline itself, which never observes them.

\subsubsection{Latent Space Analysis}
\label{subsubsec:latent}

The pseudo-group mechanism rests on a premise: that the self-supervised representation organizes SAR time series by crop type, so that naming an entire group is a meaningful operation. This subsection examines that premise directly, before turning to classification accuracy. All results reported here use the best-performing configuration in each region, 256 prototypes in CV and 128 in LEM.

\begin{figure}[H]
    \centering
    \includegraphics[draft=false, width=\linewidth]{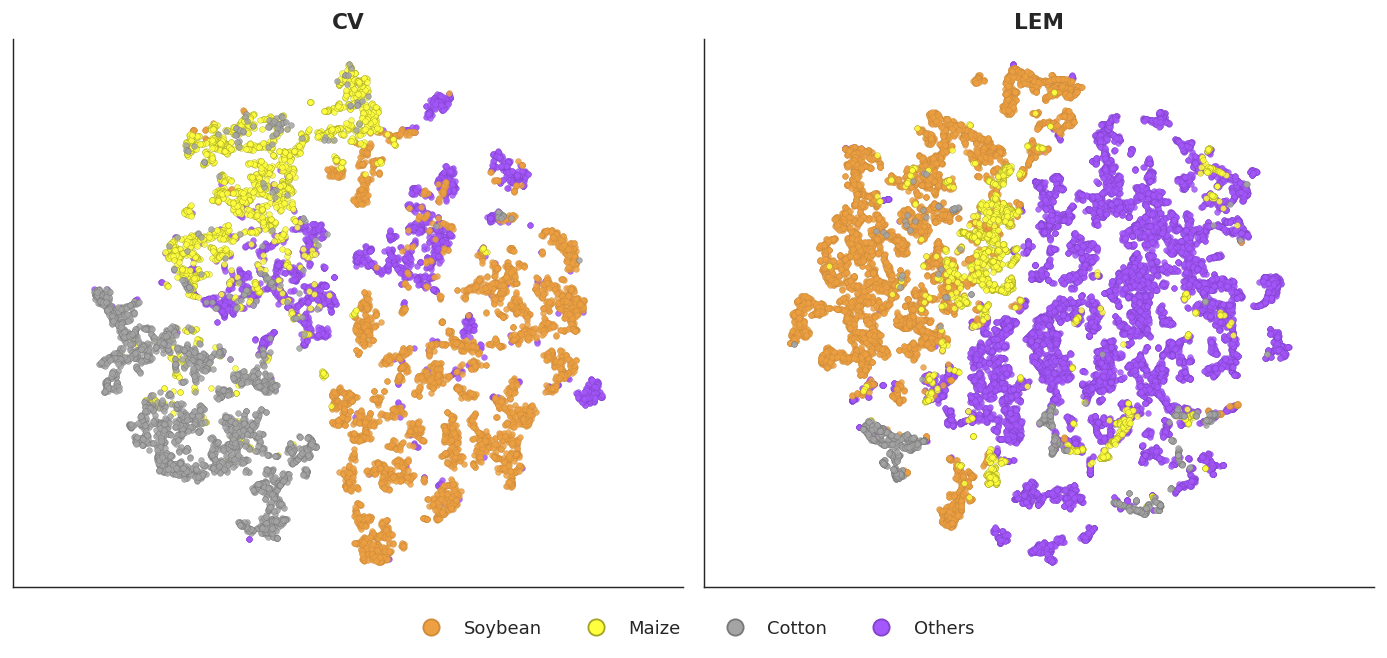}
    \caption{t-SNE projection of the MSN patch embeddings, colored by reference class.}
    \label{fig:tsne_msn}
\end{figure}

Figure~\ref{fig:tsne_msn} projects the patch embeddings onto two dimensions with t-SNE, colored by reference class. In both regions the structure induced by MSN is visibly organized by crop: the several prototypes that a single class spans appear contiguous rather than scattered, forming connected regions of a single color. In CV, cotton, maize, and soybean occupy well-defined and largely disjoint areas of the projection. In LEM, soybean and ``Others'' separate along opposite sides, with maize forming its own band between them. This organization emerges without any label being observed during training.

\begin{figure}[H]
    \centering
    \includegraphics[draft=false, width=\linewidth]{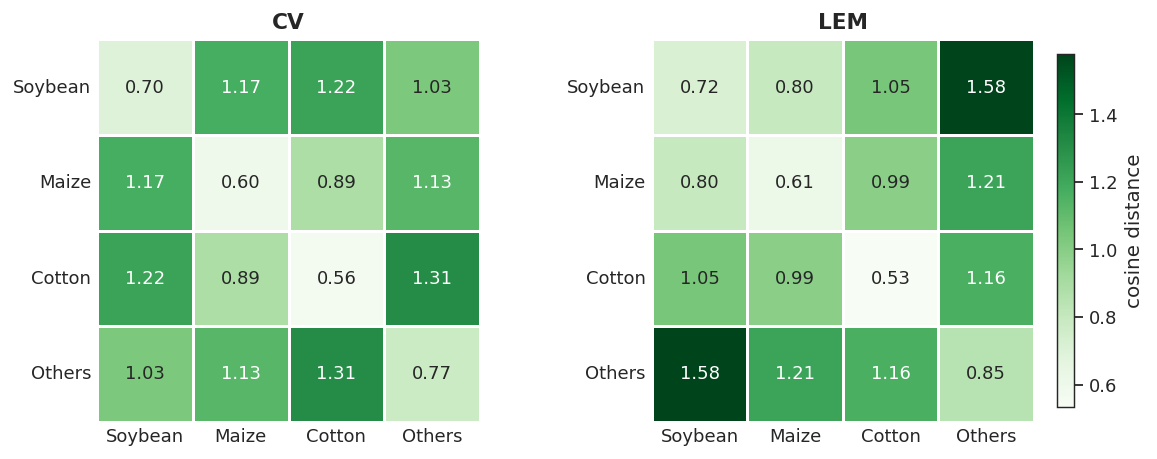}
    \caption{Mean pairwise cosine distance within and between reference classes in the MSN latent space.}
    \label{fig:distancia_msn}
\end{figure}

Because t-SNE preserves neighborhoods but not distances, Figure~\ref{fig:distancia_msn} quantifies the same structure directly in the latent space. In both regions, and for every class, the intra-class distance on the diagonal is smaller than each of the inter-class distances involving that class: patches of a given crop are closer to one another than to patches of any other crop. ``Others'' presents the largest intra-class dispersion, 0.77 in CV and 0.85 in LEM, which is expected given that it aggregates several distinct crops and land cover types under a single name.

\begin{figure}[H]
    \centering
    \includegraphics[draft=false, width=\linewidth]{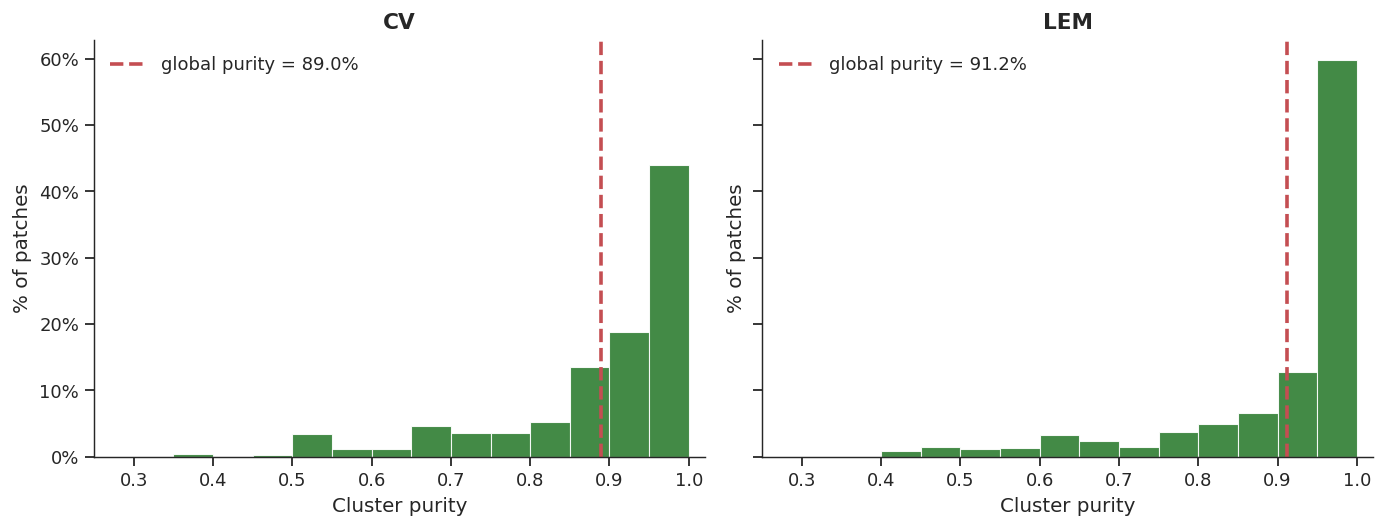}
    \caption{Distribution of cluster purity with respect to the reference classes, weighted by patch count. The dashed line marks global purity.}
    \label{fig:pureza_msn}
\end{figure}

Figure~\ref{fig:pureza_msn} reports the distribution of cluster purity, measured against the reference classes. Global purity reaches 89.0\% in CV and 91.2\% in LEM, and the distribution is strongly concentrated at the upper end, with most patches falling in clusters composed almost entirely of a single crop. This is what licenses the pseudo-group strategy: when a cluster is nearly pure, the noisy labels it contains overwhelmingly point to a single class, so the group receives the correct name and the few mislabeled patches inside it do not drive its classification. Group-level supervision is, in this sense, a consequence of the geometry of the representation rather than a heuristic convenience.

\subsubsection{Results}

Tables~\ref{tab:CV_rotulado_patches} and~\ref{tab:LEM_rotulado_patches} report model performance under different prototype configurations compared to the MapBiomas patch-level baseline. The $\Delta$ column quantifies the accuracy gain or loss relative to MapBiomas when adopting the pseudo-group strategy.

In Campo Verde (Table~\ref{tab:CV_rotulado_patches}), performance is sensitive to the number of prototypes: small configurations, such as 16 prototypes, lead to degradation below the baseline. From 32 prototypes onward, the model consistently outperforms MapBiomas. The 256-prototype configuration achieved the best results, with gains of 5.04\% in overall accuracy and 6.13\% in mean class accuracy.

\begin{figure}[H]
    \centering
    \begin{subfigure}[b]{0.48\linewidth}
        \includegraphics[draft=false, width=\linewidth]{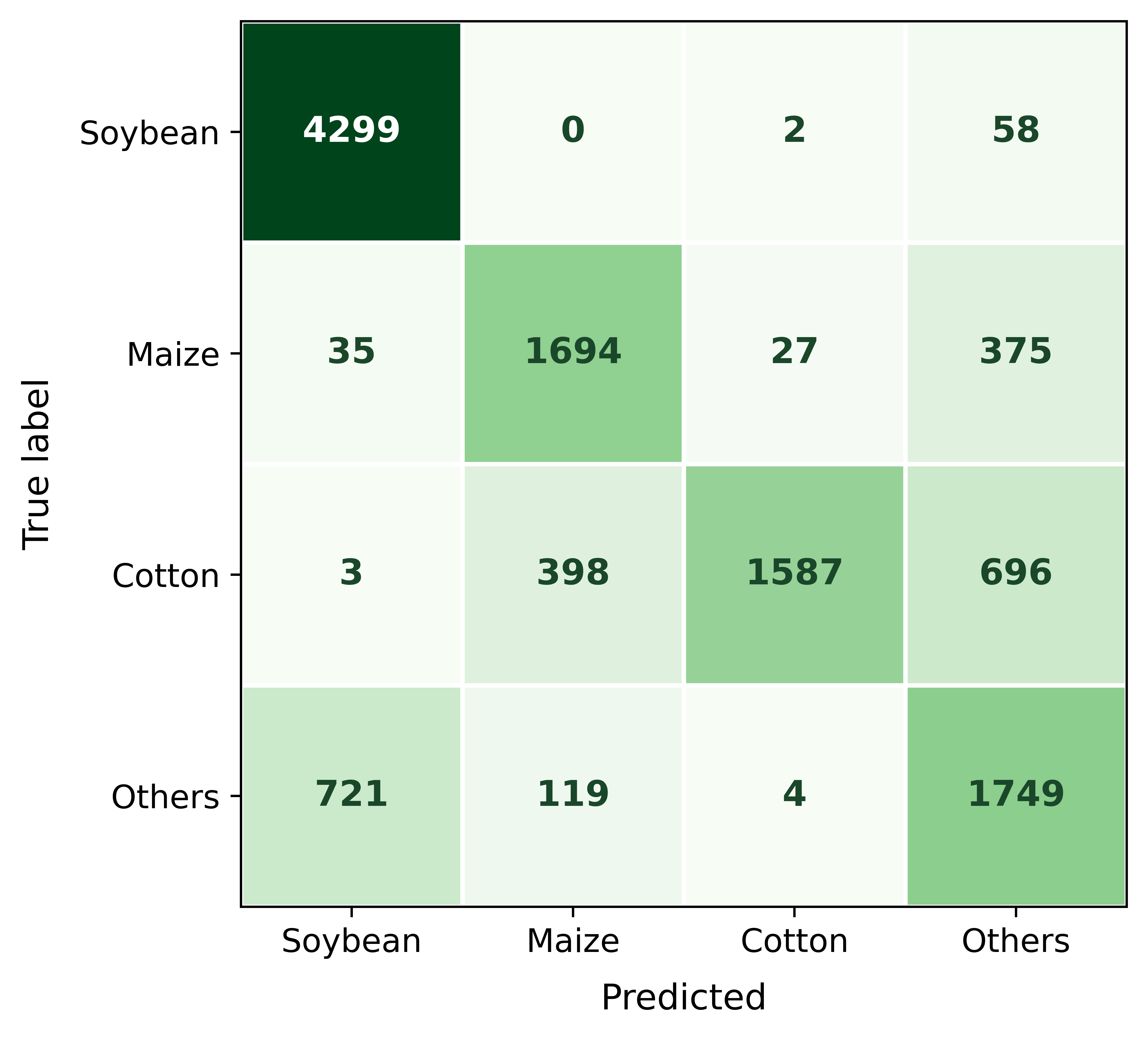}
        \caption{CV - MapBiomas - Absolute counts.}
    \end{subfigure}
    \hfill
    \begin{subfigure}[b]{0.48\linewidth}
        \includegraphics[draft=false, width=\linewidth]{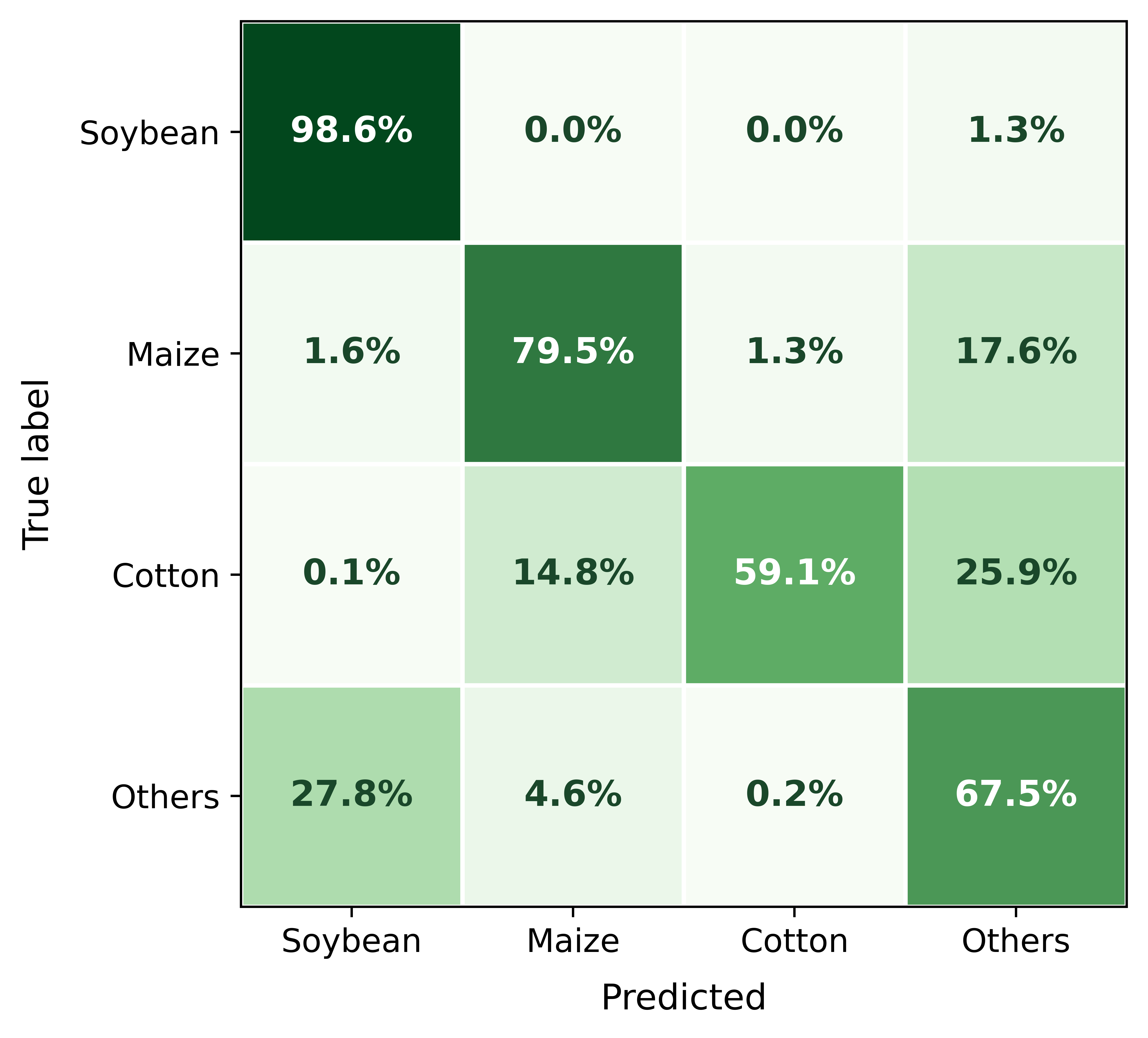}
        \caption{CV - MapBiomas - Percentage per class.}
    \end{subfigure}
    \caption{Patch-level confusion matrices for Campo Verde, MapBiomas.}
    \label{fig:CV_matriz_patches_mb}
\end{figure}

\begin{figure}[H]
    \centering
    \begin{subfigure}[b]{0.48\linewidth}
        \includegraphics[draft=false, width=\linewidth]{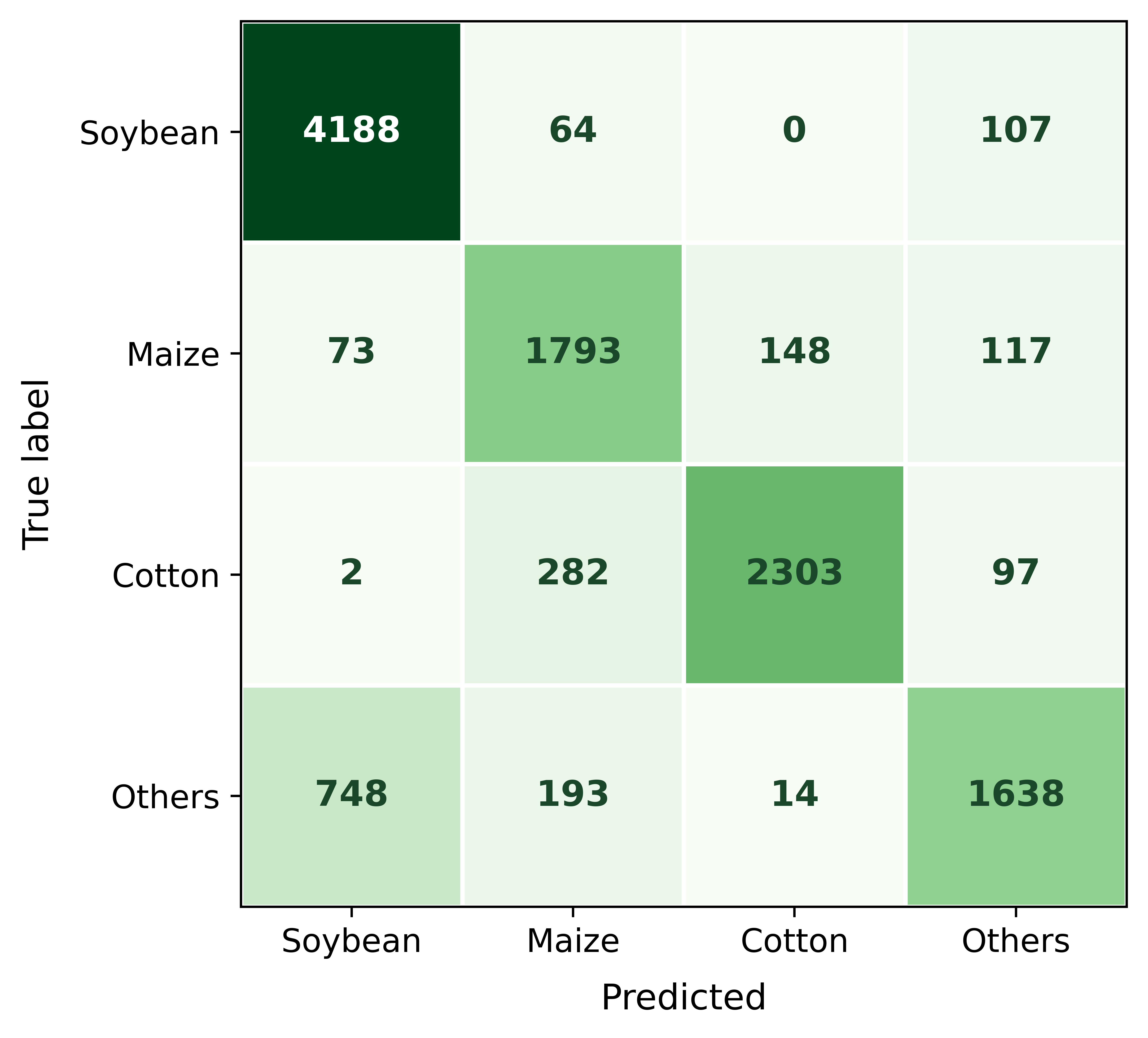}
        \caption{CV - MSN - Absolute counts.}
    \end{subfigure}
    \hfill
    \begin{subfigure}[b]{0.48\linewidth}
        \includegraphics[draft=false, width=\linewidth]{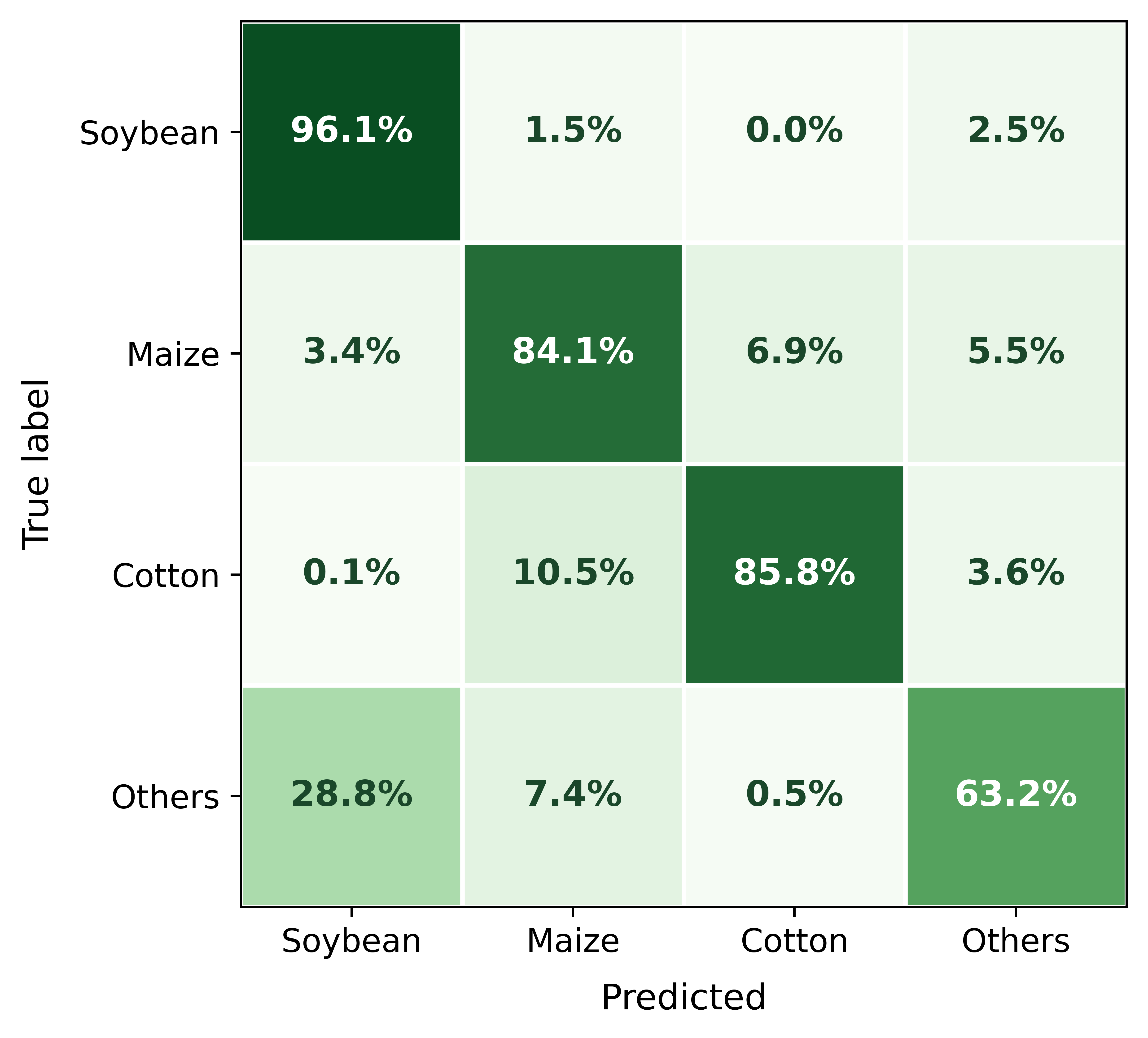}
        \caption{CV - MSN - Percentage per class.}
    \end{subfigure}
    \caption{Patch-level confusion matrices for Campo Verde, MSN with 256 prototypes.}
    \label{fig:CV_matriz_patches_msn}
\end{figure}

The confusion matrix analysis (Figures~\ref{fig:CV_matriz_patches_mb} and~\ref{fig:CV_matriz_patches_msn}) reveals the nature of this gain. MapBiomas exhibits a systematic bias of classifying maize and cotton as ``Others'', and mislabeling part of ``Others'' as soybean. The MSN model with 256 prototypes substantially reduces these errors: recall for maize and cotton improves markedly, resulting in a more balanced classification with higher mean class accuracy, without compromising the overall metric.

\begin{figure}[H]
    \centering
    \begin{subfigure}[b]{0.48\linewidth}
        \includegraphics[draft=false, width=\linewidth]{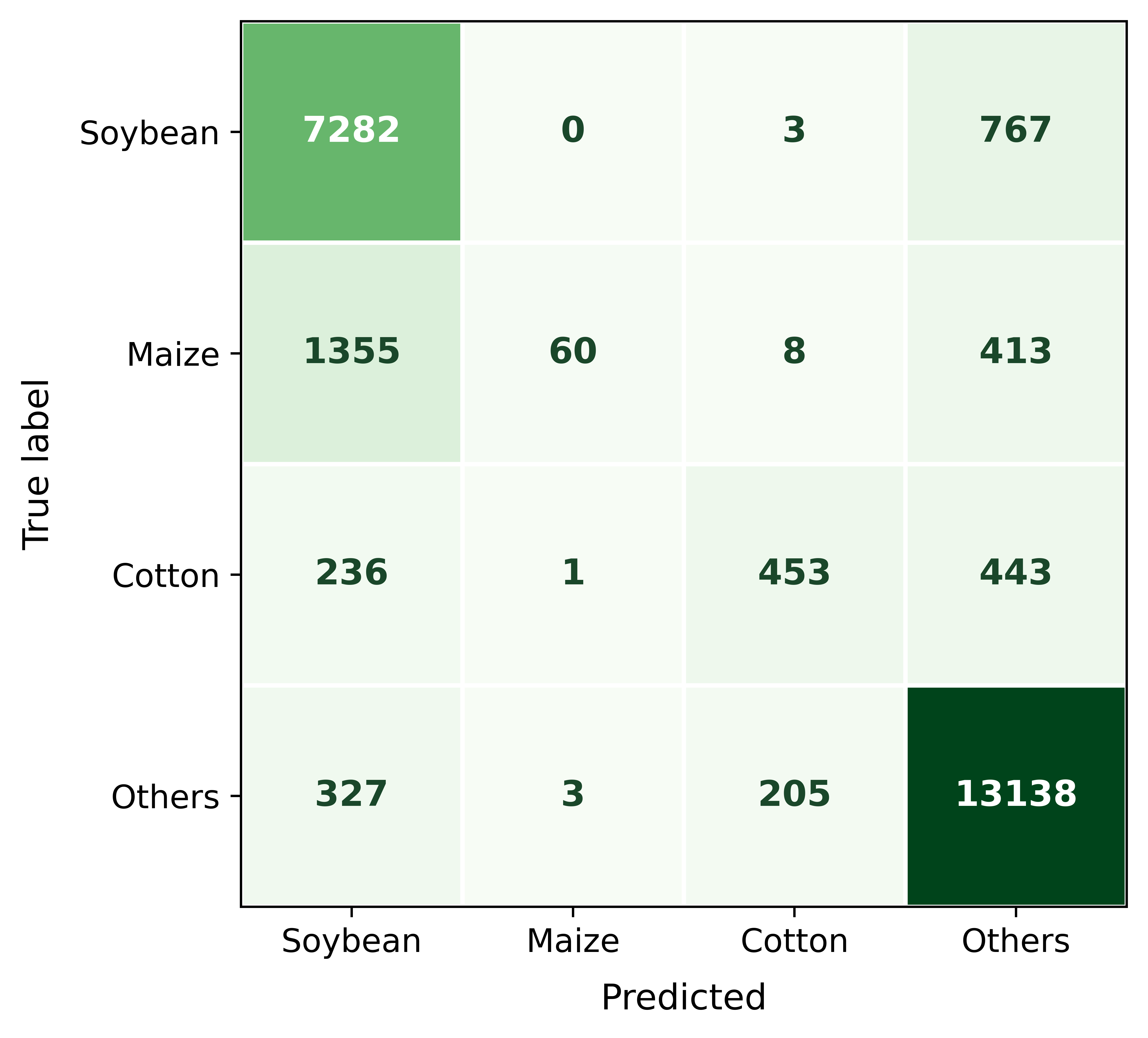}
        \caption{LEM - MapBiomas - Absolute counts.}
    \end{subfigure}
    \hfill
    \begin{subfigure}[b]{0.48\linewidth}
        \includegraphics[draft=false, width=\linewidth]{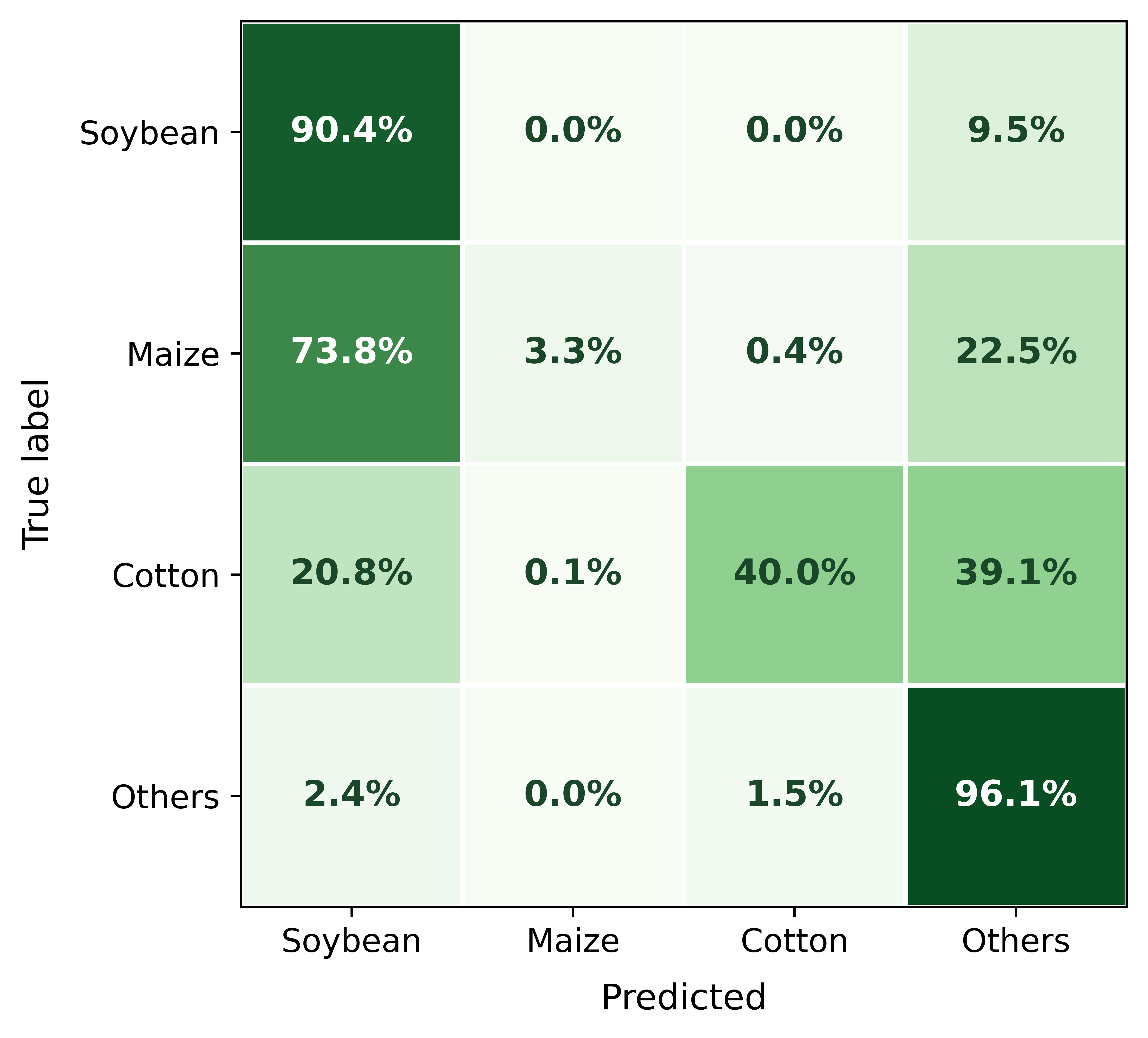}
        \caption{LEM - MapBiomas - Percentage per class.}
    \end{subfigure}
    \caption{Patch-level confusion matrices for Luís Eduardo Magalhães, MapBiomas.}
    \label{fig:LEM_matriz_patches_mb}
\end{figure}

\begin{figure}[H]
    \centering
    \begin{subfigure}[b]{0.48\linewidth}
        \includegraphics[draft=false, width=\linewidth]{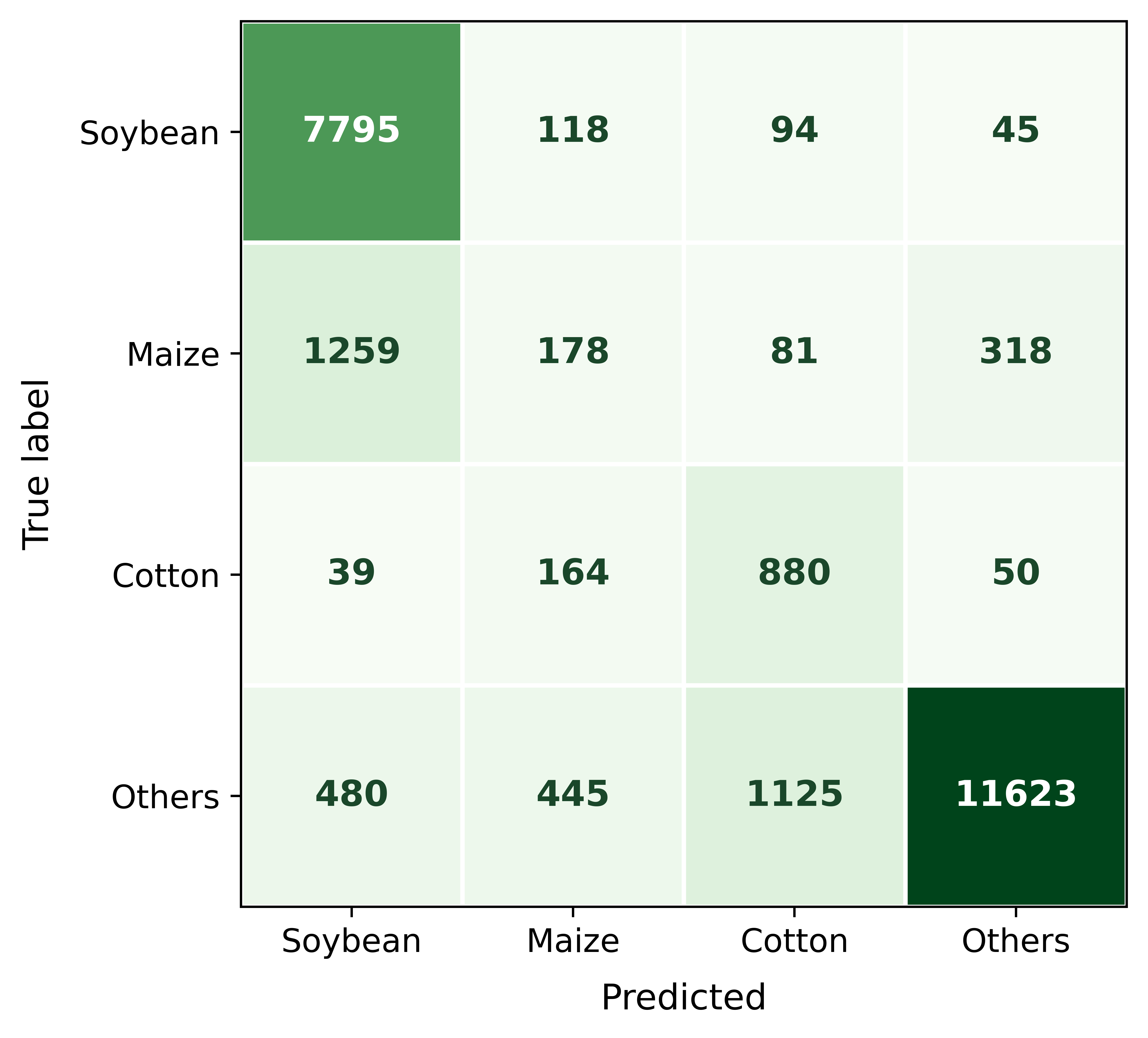}
        \caption{LEM - MSN - Absolute counts.}
    \end{subfigure}
    \hfill
    \begin{subfigure}[b]{0.48\linewidth}
        \includegraphics[draft=false, width=\linewidth]{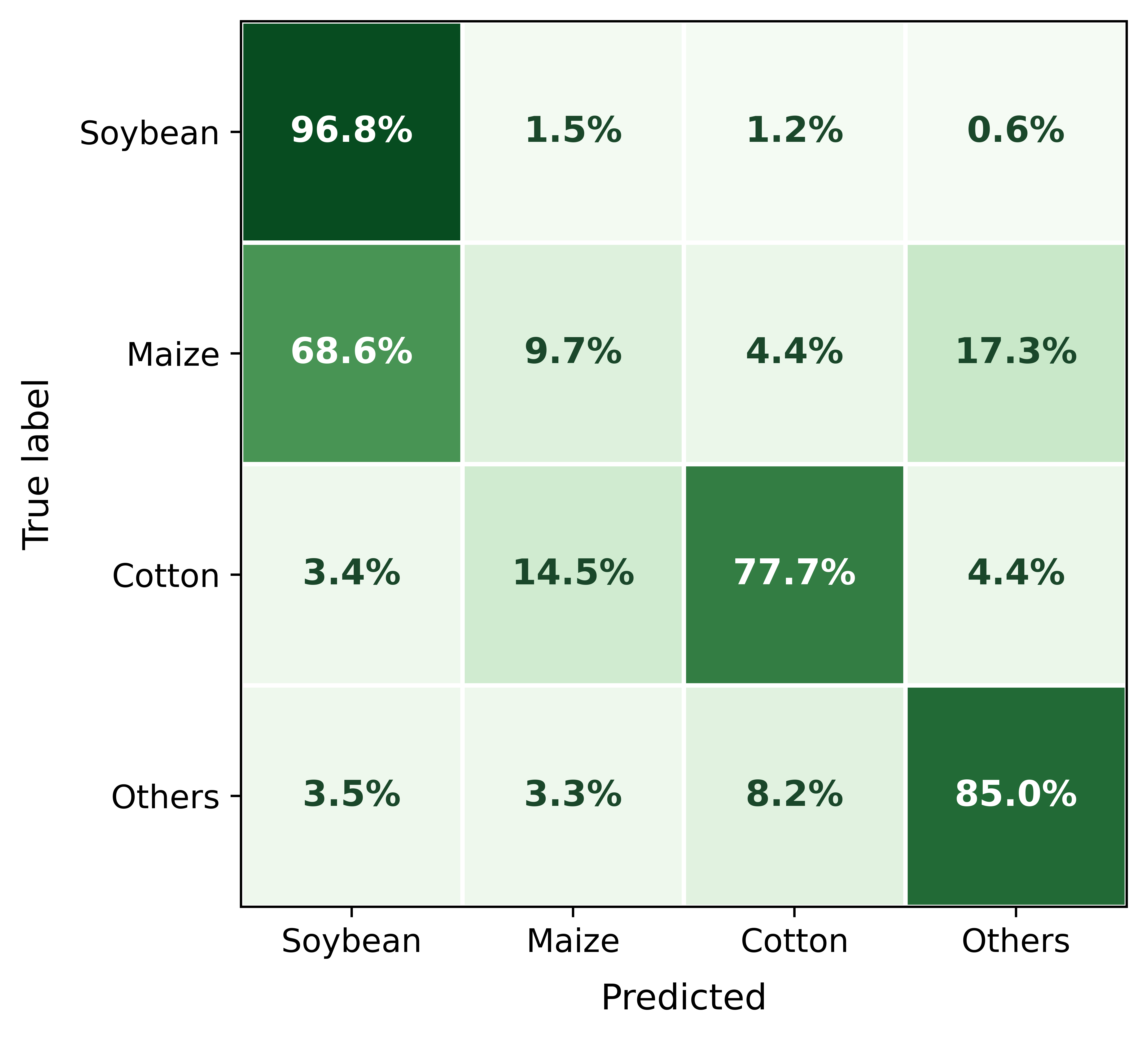}
        \caption{LEM - MSN - Percentage per class.}
    \end{subfigure}
    \caption{Patch-level confusion matrices for Luís Eduardo Magalhães, MSN with 128 prototypes.}
    \label{fig:LEM_matriz_patches_msn}
\end{figure}

In Luís Eduardo Magalhães (Table~\ref{tab:LEM_rotulado_patches}), the mean class accuracy gain is even more pronounced, surpassing MapBiomas by up to 9.86\% with 128 prototypes, albeit with a modest reduction in overall accuracy. This pattern reflects the severe limitations of MapBiomas in this region: as illustrated in Figure~\ref{fig:LEM_matriz_patches_mb}, the baseline predictions concentrate predominantly on the ``Soybean'' and ``Others'' classes, identifying only 64 maize patches against 1,836 labeled, and 669 cotton patches against 1,133. The MSN with 128 prototypes (Figure~\ref{fig:LEM_matriz_patches_msn}) substantially recovers the representativeness of these crops, even starting from strongly biased reference labels, demonstrating the capacity of the approach to mitigate the bias of the noisy label source.

The latent space analysis of Section~\ref{subsubsec:latent} accounts for the asymmetry between the two regions. Global purity is comparable in CV and LEM, 89.0\% and 91.2\%, so the self-supervised representation separates the crops equally well in both, independently of MapBiomas. The lower accuracy observed in LEM therefore does not originate in the representation but in the quality of the labels available to name the clusters: to name groups that the embedding already separates, MapBiomas offers only 64 maize patches in LEM against 2,211 in CV. Where the noisy source describes a crop poorly, the corresponding groups are harder to name correctly, even when the representation isolates them well.

\begin{table}[h!]
\centering
\small
\caption{Patch-level accuracy comparison between MSN models and the MapBiomas baseline: Campo Verde.}
\begin{tabular}{lcccc}
\hline
\textbf{Model}
& \makecell{\textbf{OA} \\ \textbf{Patches}}
& \makecell{\textbf{MA} \\ \textbf{Patches}}
& \makecell{\textbf{$\Delta$ vs MB} \\ \textbf{Overall}}
& \makecell{\textbf{$\Delta$ vs MB} \\ \textbf{Mean}} \\
\hline
MSN - 16 prototypes   & 70.17\% & 72.57\% & -9.11\%  & -3.60\% \\
MSN - 32 prototypes   & 81.86\% & 79.50\% &  2.58\%  &  3.33\% \\
MSN - 64 prototypes   & 82.99\% & 81.35\% &  3.71\%  &  5.18\% \\
MSN - 128 prototypes  & 81.02\% & 79.04\% &  1.74\%  &  2.87\% \\
MSN - 256 prototypes  & \textbf{84.32\%} & \textbf{82.30\%} & \textbf{5.04\%} & \textbf{6.13\%} \\
MSN - 512 prototypes  & 81.53\% & 79.53\% &  2.25\%  &  3.36\% \\
MSN - 1024 prototypes & 81.33\% & 79.26\% &  2.05\%  &  3.09\% \\
MapBiomas     & 79.28\% & 76.17\% &  -       &  -      \\
\hline
\end{tabular}
\label{tab:CV_rotulado_patches}
\end{table}

\begin{table}[h!]
\centering
\small
\caption{Patch-level accuracy comparison between MSN models and the MapBiomas baseline: Luís Eduardo Magalhães.}
\begin{tabular}{lcccc}
\hline
\textbf{Model}
& \makecell{\textbf{OA} \\ \textbf{Patches}}
& \makecell{\textbf{MA} \\ \textbf{Patches}}
& \makecell{\textbf{$\Delta$ vs MB} \\ \textbf{Overall}}
& \makecell{\textbf{$\Delta$ vs MB} \\ \textbf{Mean}} \\
\hline
MSN - 16 prototypes   & 64.32\% & 59.72\% & -20.45\% &  2.28\% \\
MSN - 32 prototypes   & 77.45\% & 64.31\% &  -7.32\% &  6.87\% \\
MSN - 64 prototypes   & 83.60\% & 66.81\% &  -1.17\% &  9.37\% \\
MSN - 128 prototypes  & 82.92\% & \textbf{67.30\%} & -1.85\% & \textbf{9.86\%} \\
MSN - 256 prototypes  & 83.46\% & 65.36\% &  -1.31\% &  7.92\% \\
MSN - 512 prototypes  & \textbf{83.69\%} & 64.95\% & \textbf{-1.08\%} & 7.51\% \\
MSN - 1024 prototypes & 82.80\% & 65.26\% &  -1.97\% &  7.82\% \\
MapBiomas     & 84.77\% & 57.44\% &   -      &  -      \\
\hline
\end{tabular}
\label{tab:LEM_rotulado_patches}
\end{table}

\subsection{Automatic Crop Field Mask Generation}

\subsubsection{Data}

The imagery used for automatic crop field segmentation was obtained from the Sentinel-2 collection available on the Google Earth Engine (GEE) platform. To produce a sharp, cloud-artifact-free RGB image, we computed a temporal median composite from observations collected over a seasonal interval: March to July 2016 for CV, and May to June 2020 for LEM. These intervals were selected to coincide with periods of lower cloud cover in each region, maximizing the availability of valid pixels.

Before compositing, scenes were filtered using the CloudScore+ algorithm~\cite{Pasquarella_2023_CVPR}, which identifies and excludes pixels affected by clouds and their shadows. Additionally, only scenes with cloud cover below 10\% were included. Since the goal is to produce a reference RGB image for segmentation rather than perform quantitative spectral analysis, the temporal fusion of multiple scenes does not pose a methodological concern.

\begin{figure}[H]
    \centering
    \begin{subfigure}{0.48\textwidth}
        \centering
        \includegraphics[width=\linewidth]{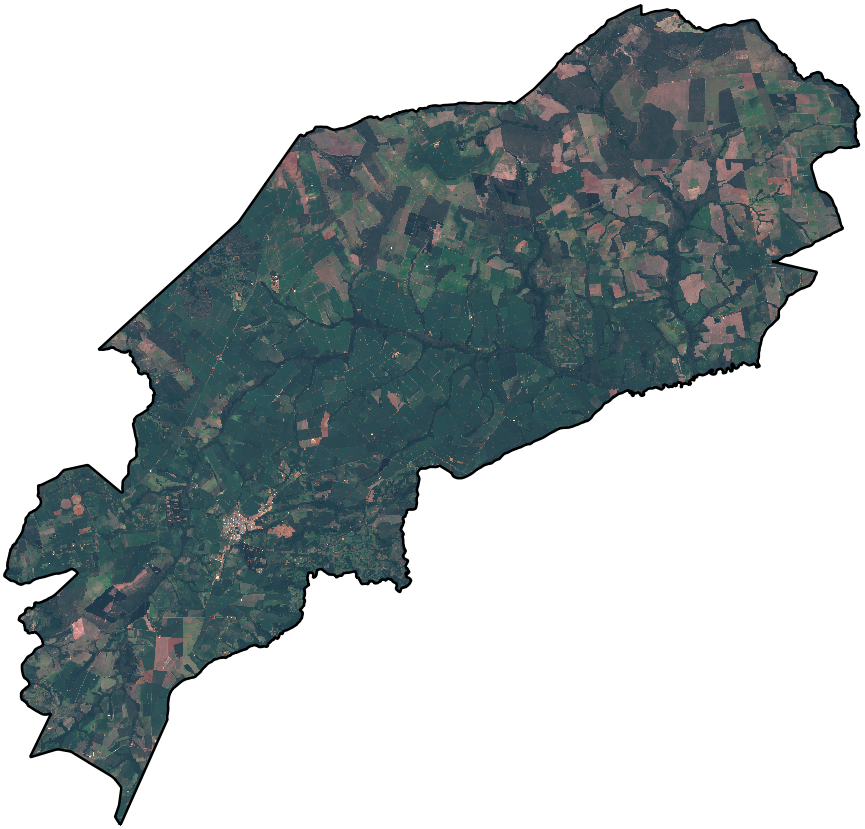}
        \caption{Campo Verde.}
    \end{subfigure}
    \hfill
    \begin{subfigure}{0.48\textwidth}
        \centering
        \includegraphics[width=\linewidth]{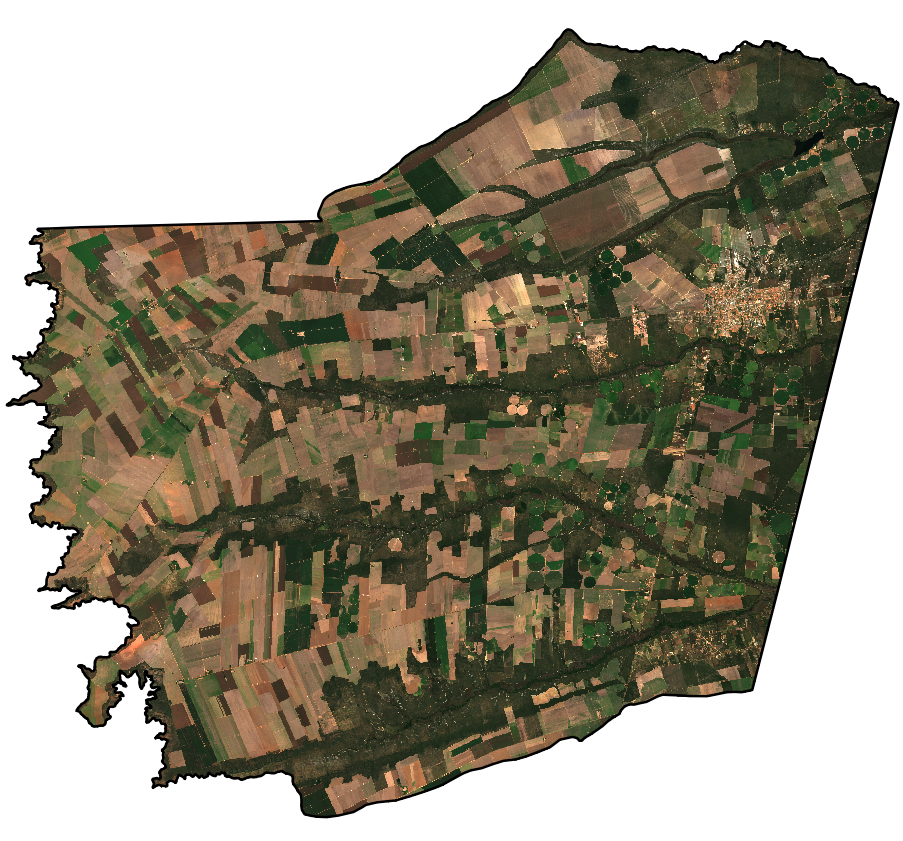}
        \caption{Luís Eduardo Magalhães.}
    \end{subfigure}
    \caption{Sentinel-2 RGB median composites used as SAM2 input: Campo Verde (Level-1C) and Luís Eduardo Magalhães (Level-2A).}
    \label{fig:areas_seg}
\end{figure}

Regarding processing level, Sentinel-2 Level-1C images (without atmospheric correction) were used for CV, due to the limited availability of Level-2A products on GEE for dates prior to 2018. For LEM, Level-2A images were available, which include atmospheric correction and yield better radiometric standardization. Figure~\ref{fig:areas_seg} shows the final RGB composites used as input to the segmentation model.

\subsubsection{Experimental Design}

Municipal images were divided into 1,024×1,024-pixel patches, a resolution compatible with the original SAM2 training, to preserve model performance during large-scale inference. To reduce discontinuities between adjacent patches arising from the independent segmentation of blocks, an overlap of 256 pixels was applied on each border. This strategy reduces edge artifacts at the cost of a proportional increase in the total number of patches processed.

Inference was performed on each patch using SAM2's automatic mask generation mode, without the need for reference points or manual annotations. Model hyperparameters were tuned to favor the detection of crop field structures, prioritizing stability and mask quality. The full configuration used is described in Table~\ref{tab:hyperparameters}.

After generation, a post-processing step was applied with two objectives: to remove regions with area below 15 hectares, a threshold below the typical minimum size of a commercial crop field, and to resolve overlaps between masks from adjacent patches introduced by the overlap strategy.

For evaluation, segmentation was initially applied only to the area covered by the reference labels, allowing direct comparison with the annotated polygons. After validation, the model was applied to the entirety of each municipality.

The nature of the problem requires an area-coverage-based evaluation rather than direct one-to-one instance matching, since the exact boundaries of crop fields predicted by SAM2 do not necessarily coincide with those in the reference dataset. In this context, Recall is the most informative metric, as it quantifies the proportion of the cultivated area effectively covered by the segmentation. The total number of predicted segments is reported as a complement, allowing the granularity of the result to be assessed.

\begin{table}[h!]
\centering
\caption{Hyperparameters used for automatic mask generation with SAM2.}
\begin{tabular}{lc}
\hline
\textbf{Parameter} & \textbf{Value} \\
\hline
points\_per\_side                 & 64            \\
points\_per\_batch                & 64            \\
pred\_iou\_thresh                 & 0.80          \\
stability\_score\_thresh          & 0.90          \\
stability\_score\_offset          & 1.0           \\
mask\_threshold                   & 0.0           \\
box\_nms\_thresh                  & 0.6           \\
crop\_n\_layers                   & 1             \\
crop\_nms\_thresh                 & 0.6           \\
crop\_overlap\_ratio              & 512/1500      \\
crop\_n\_points\_downscale\_factor & 4             \\
min\_mask\_region\_area           & 100           \\
output\_mode                      & binary\_mask  \\
use\_m2m                          & True          \\
multimask\_output                 & True          \\
\hline
\end{tabular}
\label{tab:hyperparameters}
\end{table}

\subsubsection{Results}

Table~\ref{tab:segmentation_metrics} presents the segmentation metrics obtained by SAM2 in both regions. The high Precision and Dice values reflect that the generated masks strongly overlap with the labeled areas when coverage is present. Recall, the central metric in this evaluation, reached 0.857 in CV and 0.925 in LEM, indicating that the model covered the vast majority of the cultivated area present in the images.

The number of predicted segments was remarkably close to the reference total in CV (514 predicted against 513 labeled). In LEM, the model generated fewer segments (821 against 1,069), which is expected given that several adjacent crop fields in this region present subtle boundaries in the RGB image, leading the model to merge distinct polygons into a single mask. The lower Recall in CV relative to LEM can be attributed to the use of Level-1C images, without atmospheric correction, which reduces the spectral sharpness of crop field boundaries and marginally hinders segmentation.

\begin{table}[h!]
\centering
\caption{Segmentation performance metrics for the SAM2 automatic mask generation.}
\begin{tabular}{lcccccc}
\hline
\textbf{Study Area} & \textbf{IoU} & \textbf{Dice} & \textbf{Precision} & \textbf{Recall} & \textbf{GT Masks} & \textbf{Predicted Masks} \\
\hline
CV & 0.822 & 0.902 & 0.953 & 0.857 & 513  & 514 \\
LEM         & 0.865 & 0.928 & 0.930 & 0.925 & 1069 & 821 \\
\hline
\end{tabular}
\label{tab:segmentation_metrics}
\end{table}

\begin{figure}[H]
    \centering
    \begin{subfigure}{0.48\textwidth}
        \centering
        \includegraphics[width=\linewidth]{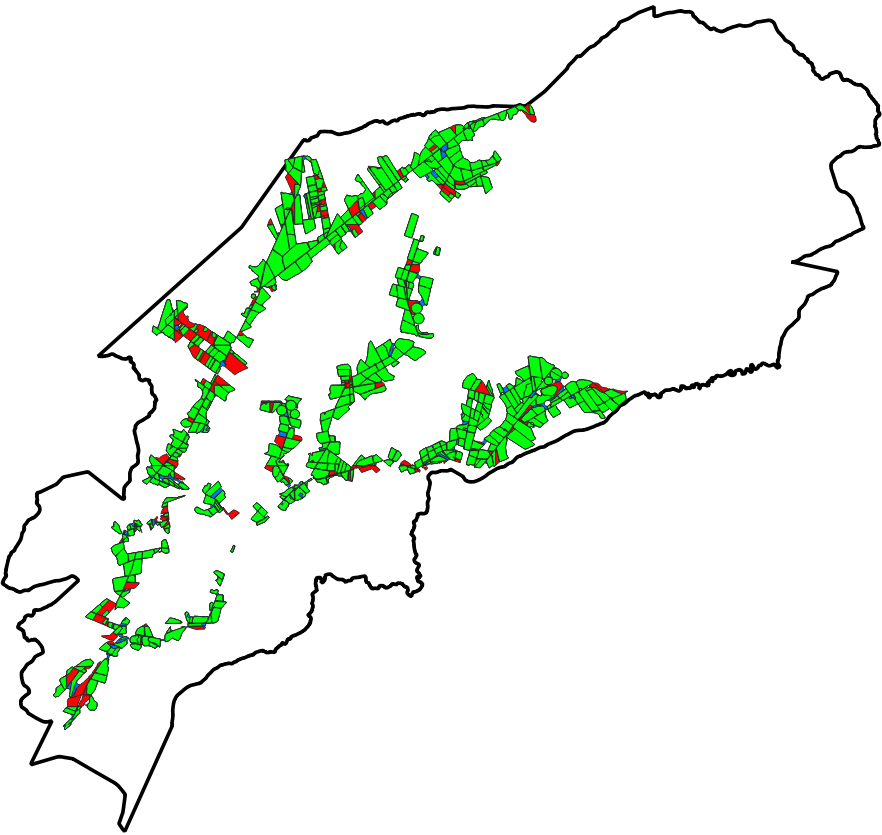}
        \caption{Campo Verde.}
    \end{subfigure}
    \hfill
    \begin{subfigure}{0.48\textwidth}
        \centering
        \includegraphics[width=\linewidth]{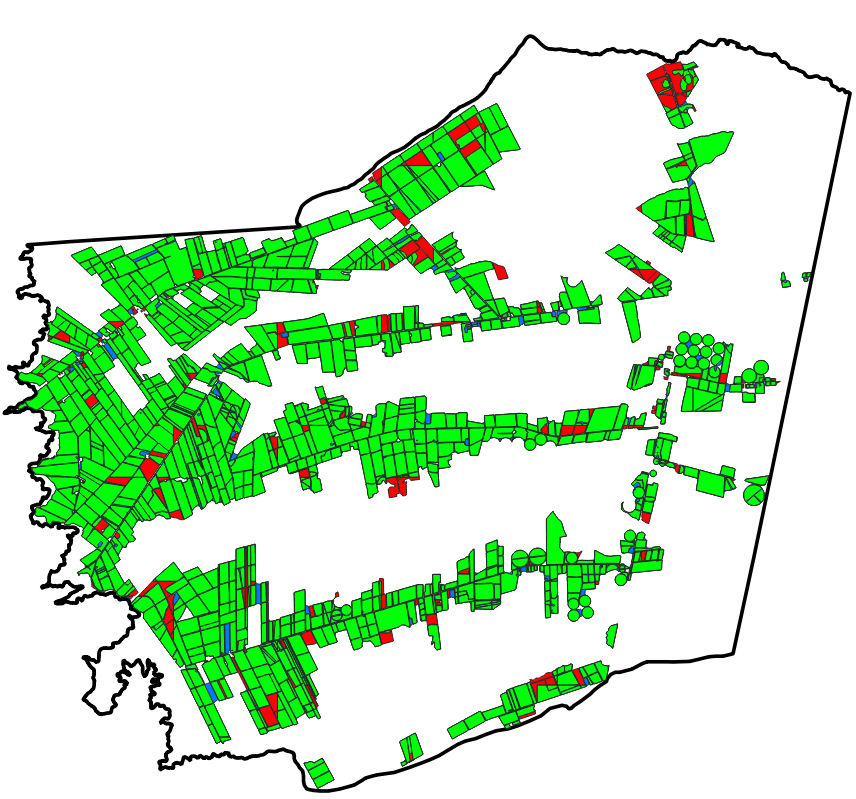}
        \caption{Luís Eduardo Magalhães.}
    \end{subfigure}
    \caption{SAM2 segmentation validation over the labeled area. Green: true positive; red: false negative; blue: prediction outside annotated boundaries.}
    \label{fig:viz_seg_rotulada}
\end{figure}

\begin{figure}[H]
    \centering
    \begin{subfigure}{0.48\textwidth}
        \centering
        \includegraphics[width=\linewidth]{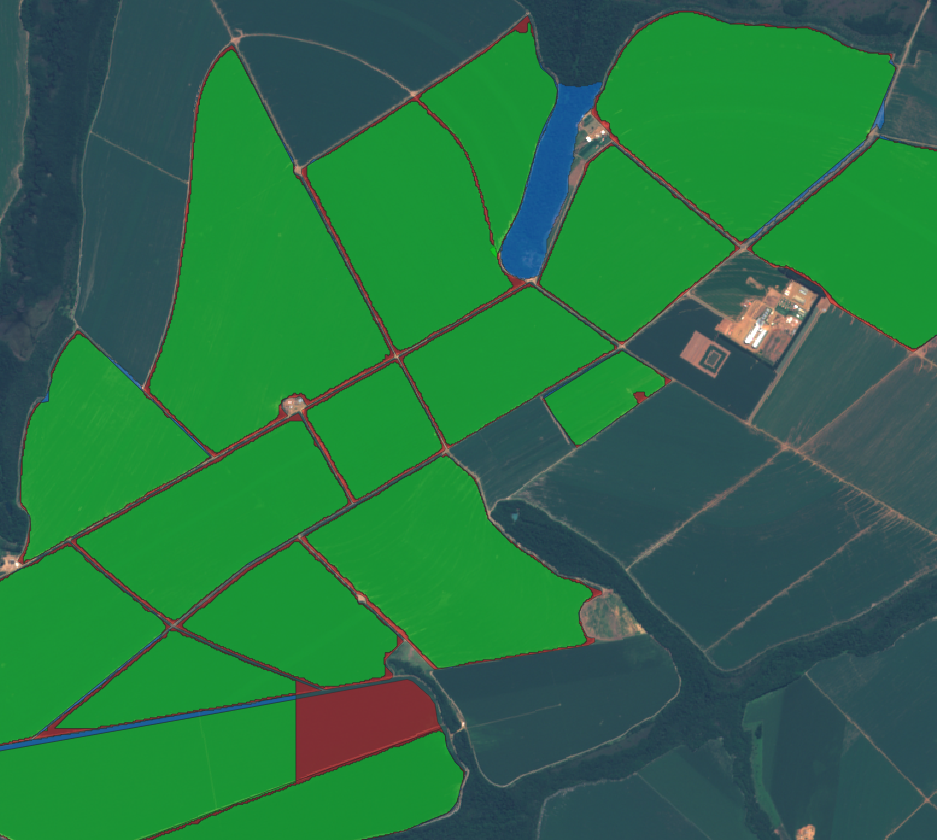}
        \caption{Campo Verde.}
    \end{subfigure}
    \hfill
    \begin{subfigure}{0.48\textwidth}
        \centering
        \includegraphics[width=\linewidth]{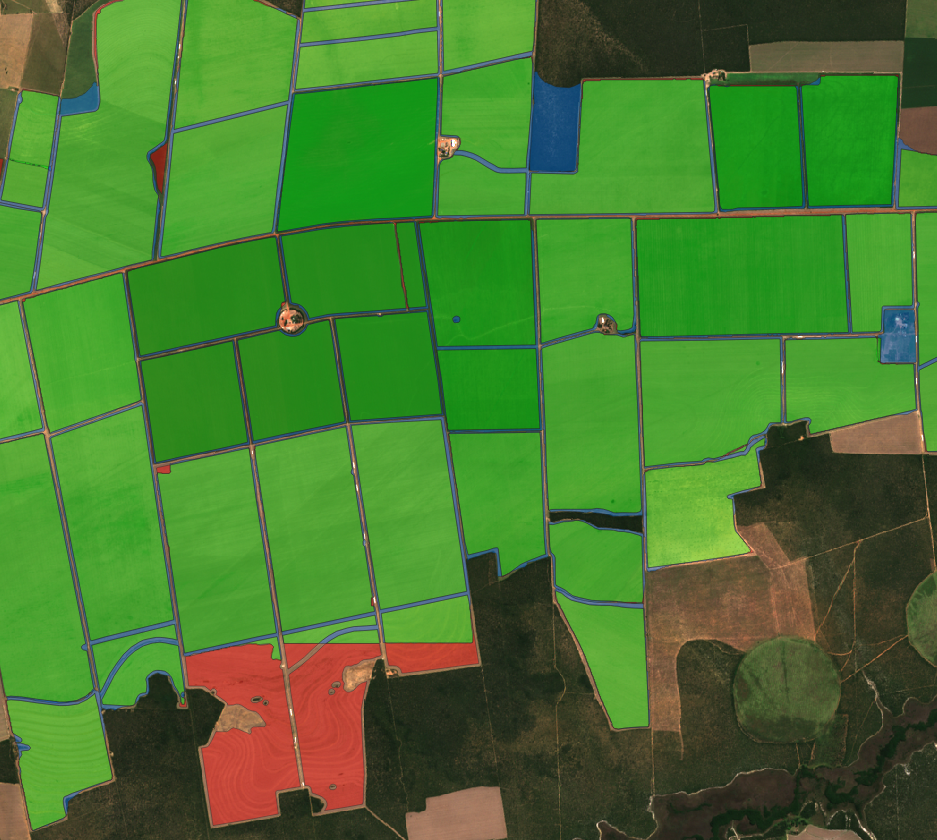}
        \caption{Luís Eduardo Magalhães.}
    \end{subfigure}
    \caption{Close-up of SAM2 segmentation over the labeled area.}
    \label{fig:viz_seg_rotulada_zoom}
\end{figure}

Figure~\ref{fig:viz_seg_rotulada} illustrates the results over the labeled area: green regions indicate the intersection between the label and the prediction; red regions indicate labeled area not covered by the model (false negative); and blue regions indicate area predicted outside the reference label boundaries. The predominance of green confirms that the majority of the cultivated area was correctly segmented. Figure~\ref{fig:viz_seg_rotulada_zoom} details these boundaries, showing that the predicted limits closely follow the crop field borders.

\begin{figure}[H]
    \centering
    \begin{subfigure}{0.48\textwidth}
        \centering
        \includegraphics[width=\linewidth]{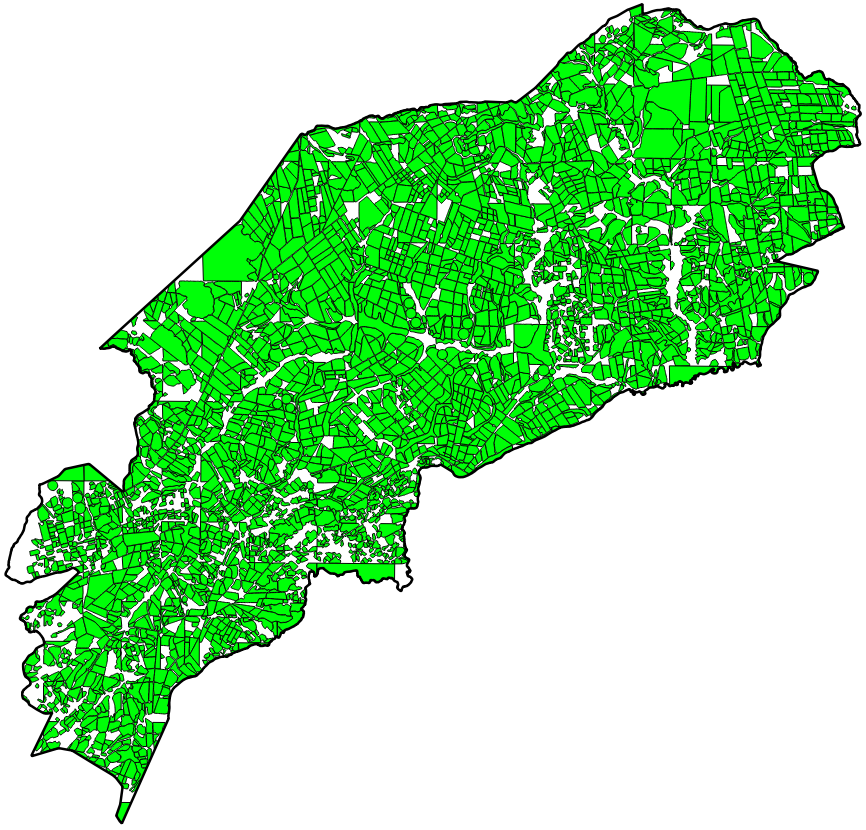}
        \caption{Campo Verde: 3,758.08~km² segmented.}
    \end{subfigure}
    \hfill
    \begin{subfigure}{0.48\textwidth}
        \centering
        \includegraphics[width=\linewidth]{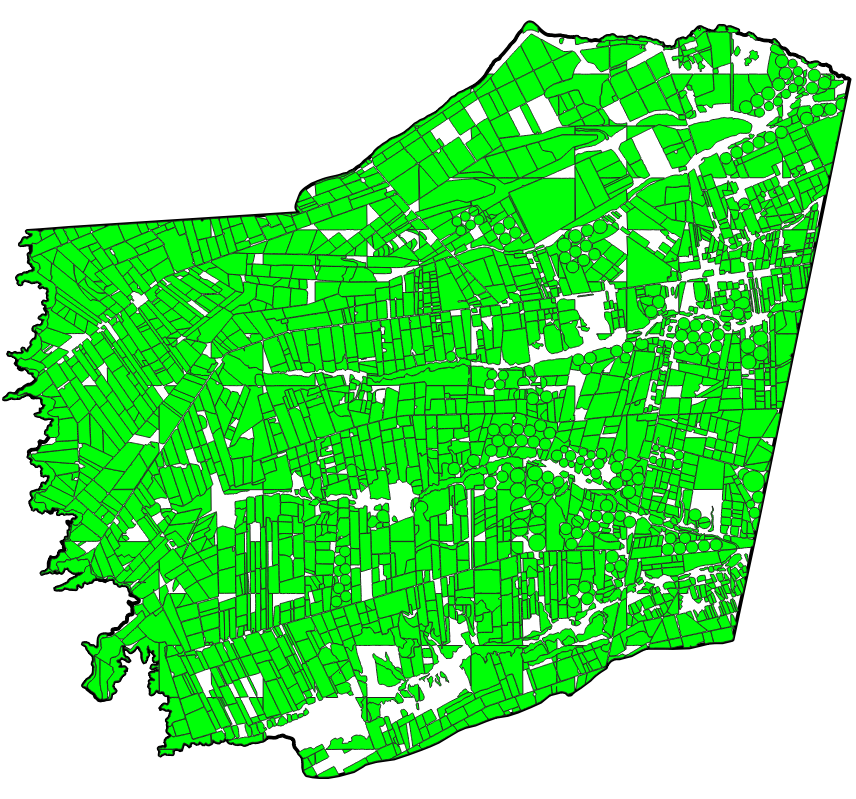}
        \caption{Luís Eduardo Magalhães: 3,344.19~km² segmented.}
    \end{subfigure}
    \caption{SAM2 segmentation applied to the full municipality.}
    \label{fig:viz_seg_full}
\end{figure}

\begin{figure}[H]
    \centering
    \begin{subfigure}{0.48\textwidth}
        \centering
        \includegraphics[width=\linewidth]{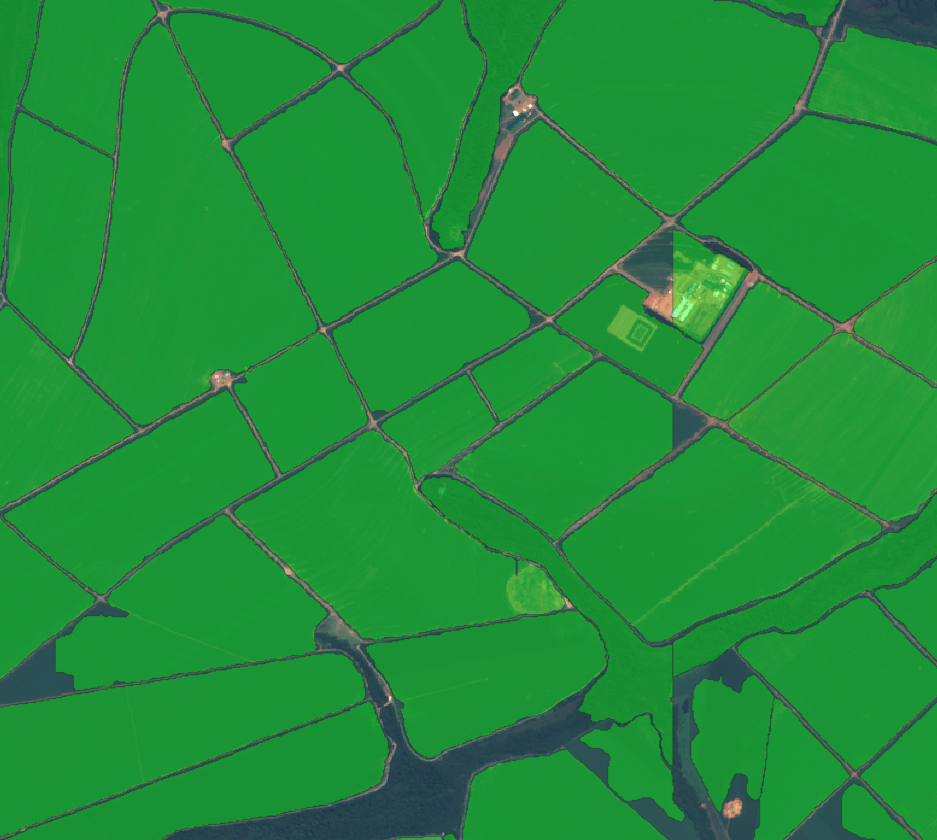}
        \caption{Campo Verde.}
    \end{subfigure}
    \hfill
    \begin{subfigure}{0.48\textwidth}
        \centering
        \includegraphics[width=\linewidth]{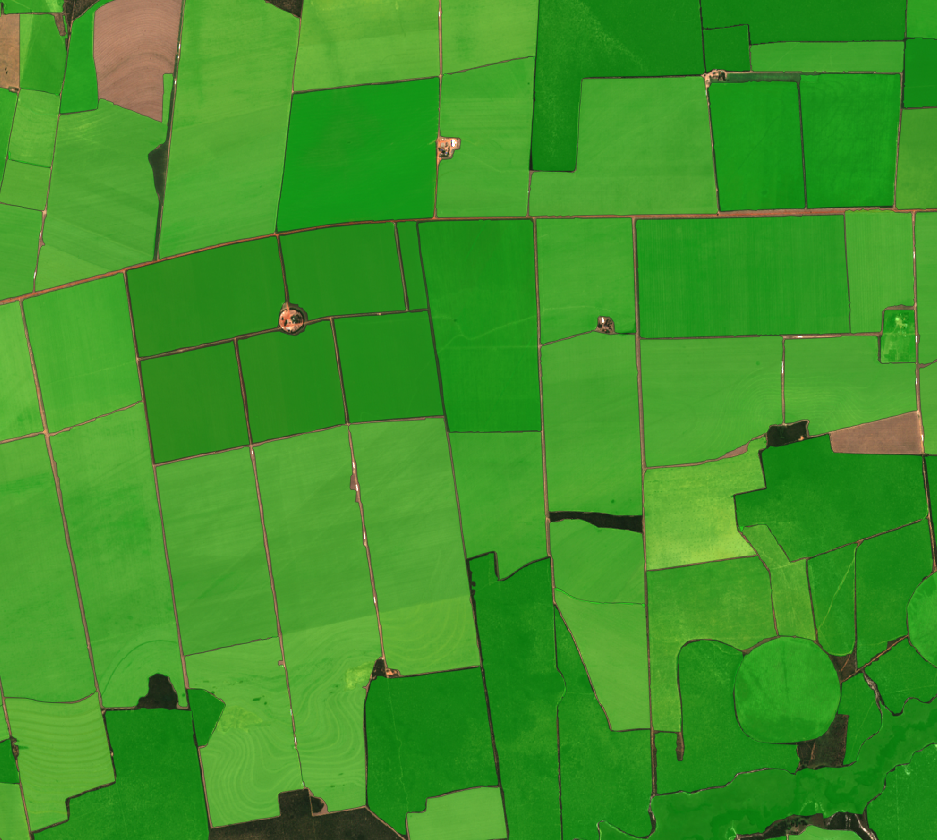}
        \caption{Luís Eduardo Magalhães.}
    \end{subfigure}
    \caption{Close-up of SAM2 segmentation at the municipal scale.}
    \label{fig:viz_seg_full_zoom}
\end{figure}

Applied to the full municipality, SAM2 substantially expanded the segmented area: in CV, from 628.72~km² to 3,758.08~km² (approximately 6× increase); in LEM, from 1,276.59~km² to 3,344.19~km² (approximately 2.6× increase). These results extend beyond agricultural crop fields, also including urban areas, forest fragments, and water bodies visible in the composites, reflecting the model’s generalization capacity. Figure~\ref{fig:viz_seg_full} shows the resulting coverage at the municipal scale, and Figure~\ref{fig:viz_seg_full_zoom} displays the level of detail achieved at crop field boundaries.

Overall, SAM2 proved to be an efficient and scalable approach for the automatic segmentation of agricultural crop fields from satellite RGB imagery. The results show that the model, without any fine-tuning for the agricultural domain and without a single manually drawn polygon, is able to delineate crop field structures with sufficient quality to enable the subsequent classification step. By furnishing the field geometries that anchor the spatial-consistency mechanism, this automatic delineation is what allows intra-field noise to be suppressed over arbitrarily large areas: the capacity to expand segmented crop field coverage by multiple factors, with spatial consistency and without any additional annotation, is precisely what makes the proposed methodology applicable at the municipal and state scales.


\subsection{Crop Field Classification}

\subsubsection{Data}

This step integrates the outputs of the two previous stages: the MSN prototypes trained on Sentinel-1 time series and the crop field polygons automatically generated by SAM2. For evaluation over the labeled area, the annotated polygons from CV and LEM are used as reference units. For evaluation at the municipal scale, the crop fields automatically segmented by SAM2 over the entirety of each municipality are used.

For the state-scale generalization analysis, the methodology was applied to the entire state of Mato Grosso and to the agricultural region of Bahia, both for 2020. The year 2016 was not used for Mato Grosso due to quality issues in the Sentinel-1 images available for that period. In this analysis, only crop fields classified by MapBiomas as temporary cropland were retained, reducing computational cost without compromising the relevance of the results for the agricultural context.

The same state-scale inference is additionally confronted with the official agricultural statistics of Brazil, taken from the Municipal Agricultural Production survey (\textit{Produção Agrícola Municipal}, PAM), published annually by the Brazilian Institute of Geography and Statistics (IBGE)~\cite{ibge_pam2020}. PAM is the official reference for crop production in the country and reports, for each of the 5,570 Brazilian municipalities, the planted area, the harvested area, the quantity produced, the average yield, and the value of production of each crop. The municipality is the finest spatial unit at which official Brazilian crop statistics are published, and it is the unit on which agricultural credit, insurance, and supply policies are actually formulated.

PAM is not a measurement of the territory, and this is essential to how the comparison should be read. Its figures are produced by state and municipal panels of the Agricultural Statistics Coordination Group (GCEA), which bring together IBGE staff and local informants from rural extension agencies, agriculture secretariats, cooperatives, and trading companies. Each municipal figure is the consensus estimate of such a panel, refined month by month over the crop year, rather than a field measurement or an estimate derived from a probabilistic sample of farms. The resulting product is therefore subject to error of a nature entirely different from that of a remote sensing product: its quality depends on the availability and expertise of local informants, revision effort is uneven across municipalities, and the area is attributed to the municipality declared by the informant, which need not be the one where the field physically lies. Small municipalities, minority crops, and second-crop areas are, for these reasons, the entries most exposed to error.

PAM is consequently not treated here as ground truth, and the comparison reported at the end of this section is not an accuracy evaluation. It is a consistency check against an independent source: the model observes satellite imagery and public land cover labels, PAM observes local declarations, and the two estimates are built from disjoint evidence. A reasonable correlation between them, particularly in the municipalities where a crop occupies a large area and both estimates are most reliable, is evidence that the pipeline produces quantities of the correct magnitude at the aggregation level at which agricultural statistics are used in practice.

\subsubsection{Experimental Design}

For each crop field, all patches within its area are classified by the MSN prototypes, and the most frequent crop among the individual predictions is assigned to the field as a whole. This majority voting reduces the influence of pointwise errors and exploits the internal consistency of the agricultural field unit, as described in Section~\ref{sec:methodology}.

Evaluation over the labeled area uses OA and MA computed at the crop field level, weighted by the area of each polygon. Two comparative deltas are reported: $\Delta$ Gain, which measures the accuracy improvement obtained by transitioning from the patch level to the crop field level, and $\Delta$ vs MB, which compares each MSN configuration with the MapBiomas baseline under the same aggregation logic.

For the state-scale analysis, the first crop and second crop of each municipality were classified and compared with MapBiomas predictions through per-class discordance, defined as $1 - \text{OA}$. The objective is not to minimize this discordance in absolute terms, since the limitations of MapBiomas as a reference have already been established; rather, the goal is low variance across municipalities, which would indicate that the methodology's behavior remains stable and predictable regardless of local particularities. To mitigate the effect of small samples, only municipalities with more than 500 segmented crop fields were included, and each class was evaluated only in municipalities with more than 35 crop fields of that category.

The comparison with IBGE/PAM operates on the same state-scale inference, aggregated by municipality, and covers soybean, maize, and cotton. The reference quantity is the harvested area in hectares reported by PAM. For each municipality and each crop, the corresponding estimate is the sum of the areas of all crop fields assigned to that crop, computed from the field geometry reprojected to the corresponding UTM zone. A field contributes once for each season in which it was classified as the crop, so a parcel planted with soybean in the first season and maize in the second contributes its area to both crops. This reproduces the way PAM accounts for area under double-cropping systems, in which the same physical area is counted once per cultivation within the crop year. Unlike the discordance analysis described above, no minimum crop field count per municipality is imposed here: every municipality with segmented crop fields enters the comparison, 122 in Mato Grosso and 35 in Bahia.

Because neither source measures the territory directly, the comparison is read as the agreement between two independent estimates, and not as the error of one against the other. It is summarized by the correlation between the two areas in hectares, with one point per municipality; the coefficient is symmetric in the two sources, so that neither is designated as the target of the other, and the logarithmic scale used in the figure below is a display choice that does not affect it. No error metric is reported, since an error would presuppose a correct value that neither source provides.

\subsubsection{Results}

Tables~\ref{tab:CV_rotulado_mask} and~\ref{tab:LEM_rotulado_mask} report model performance after crop field-level prediction aggregation. The $\Delta$ Gain column quantifies the benefit of transitioning from the patch level to the crop field level, while $\Delta$ vs MB compares each model with the MapBiomas baseline under the same aggregation logic.

\begin{figure}[H]
    \centering
    \begin{subfigure}[b]{0.48\linewidth}
        \includegraphics[draft=false, width=\linewidth]{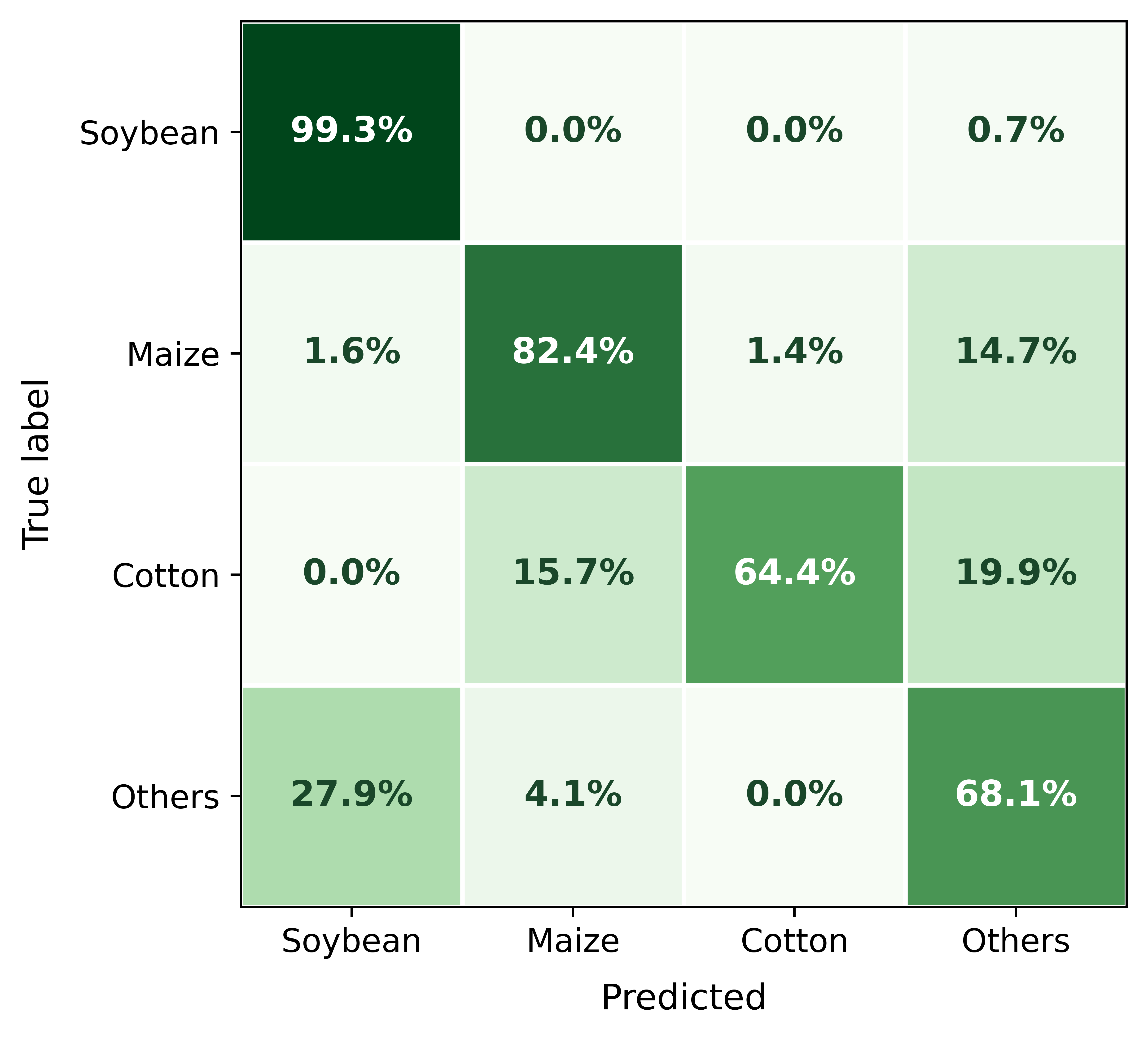}
        \caption{CV -- MapBiomas.}
    \end{subfigure}
    \hfill
    \begin{subfigure}[b]{0.48\linewidth}
        \includegraphics[draft=false, width=\linewidth]{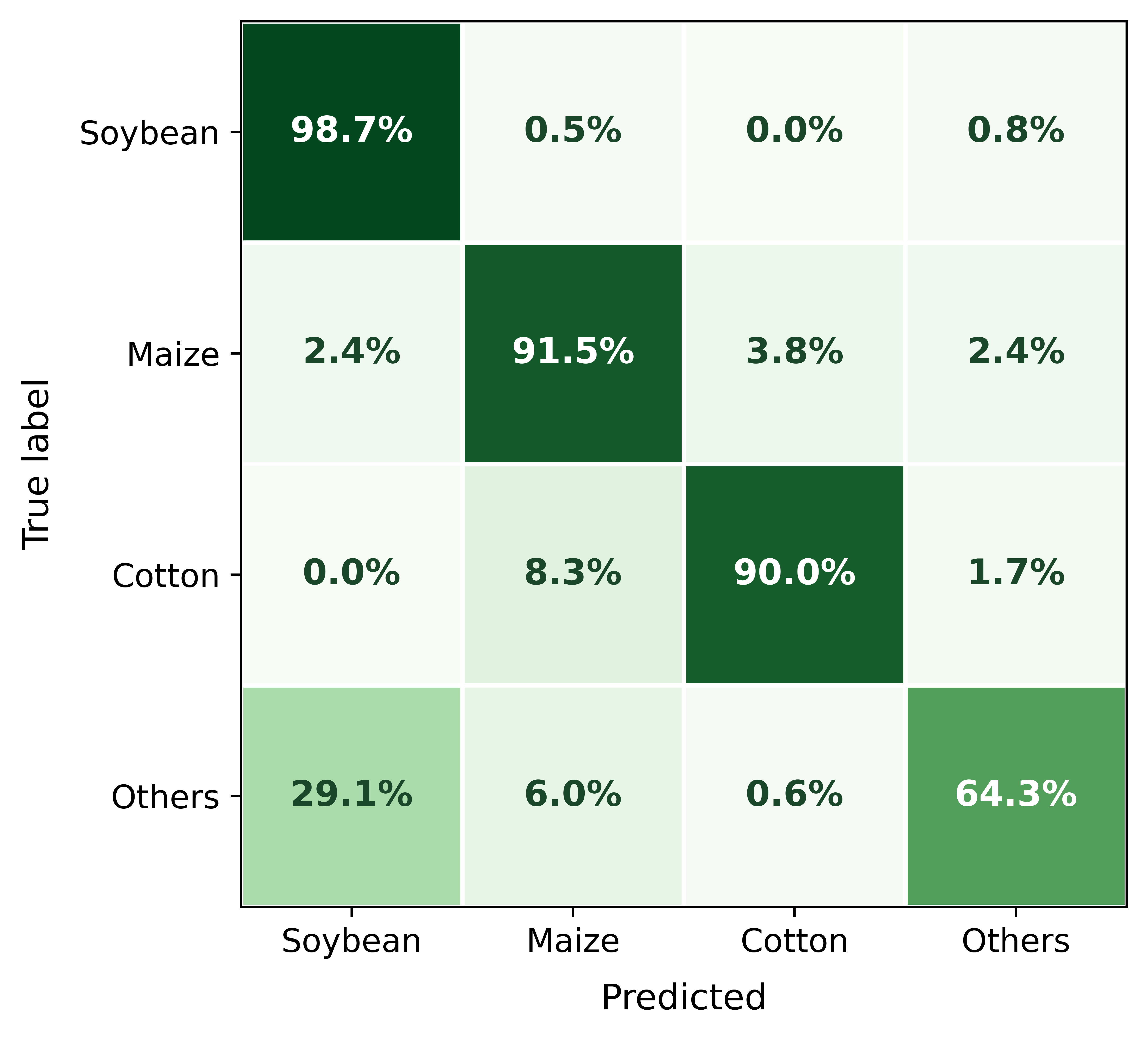}
        \caption{CV -- MSN (256 prototypes).}
    \end{subfigure}
    \caption{Area-based confusion matrices at field level for Campo Verde.}
    \label{fig:CV_matriz_area}
\end{figure}

In Campo Verde (Table~\ref{tab:CV_rotulado_mask}), spatial aggregation produced gains across all MSN configurations. The 256-prototype model achieved the best results relative to MapBiomas, with improvements of 6.45\% in overall accuracy and 7.58\% in mean class accuracy. The confusion matrices in Figure~\ref{fig:CV_matriz_area} make the nature of this gain visible: MapBiomas correctly classifies only 64.4\% of the cotton area in CV, while MSN 256 raises this to 90.0\%, a gain of 25.6 percentage points. The maize recall also improves, from 82.4\% to 91.5\%. The improvement pattern is consistent with that observed at the patch level, but amplified by the spatial smoothing effect of aggregation.

\begin{figure}[H]
    \centering
    \begin{subfigure}[b]{0.48\linewidth}
        \includegraphics[draft=false, width=\linewidth]{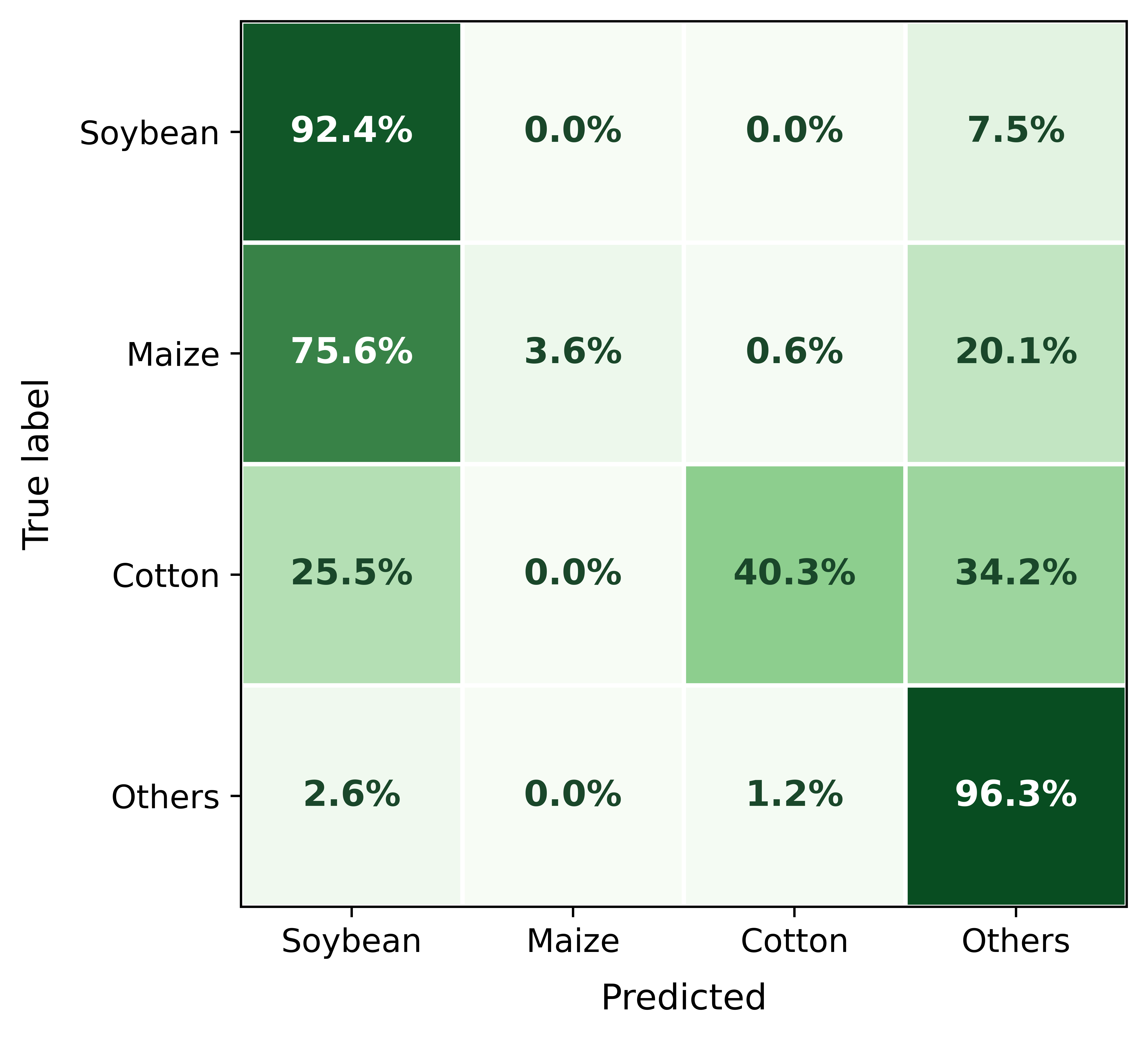}
        \caption{LEM -- MapBiomas.}
    \end{subfigure}
    \hfill
    \begin{subfigure}[b]{0.48\linewidth}
        \includegraphics[draft=false, width=\linewidth]{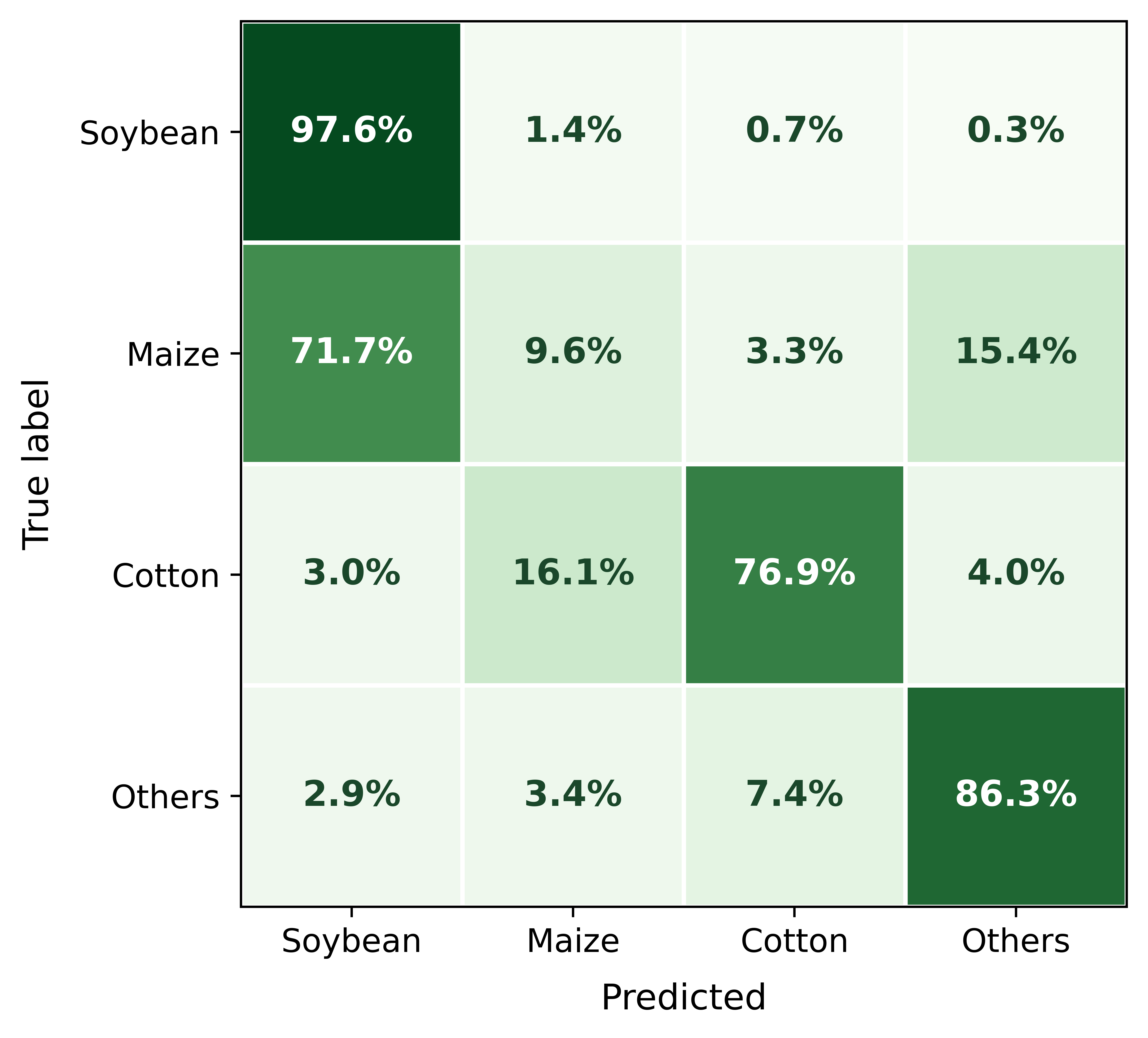}
        \caption{LEM -- MSN (128 prototypes).}
    \end{subfigure}
    \caption{Area-based confusion matrices at field level for Luís Eduardo Magalhães.}
    \label{fig:LEM_matriz_area}
\end{figure}

\begin{figure}[H]
    \centering
    \includegraphics[draft=false, width=\linewidth]{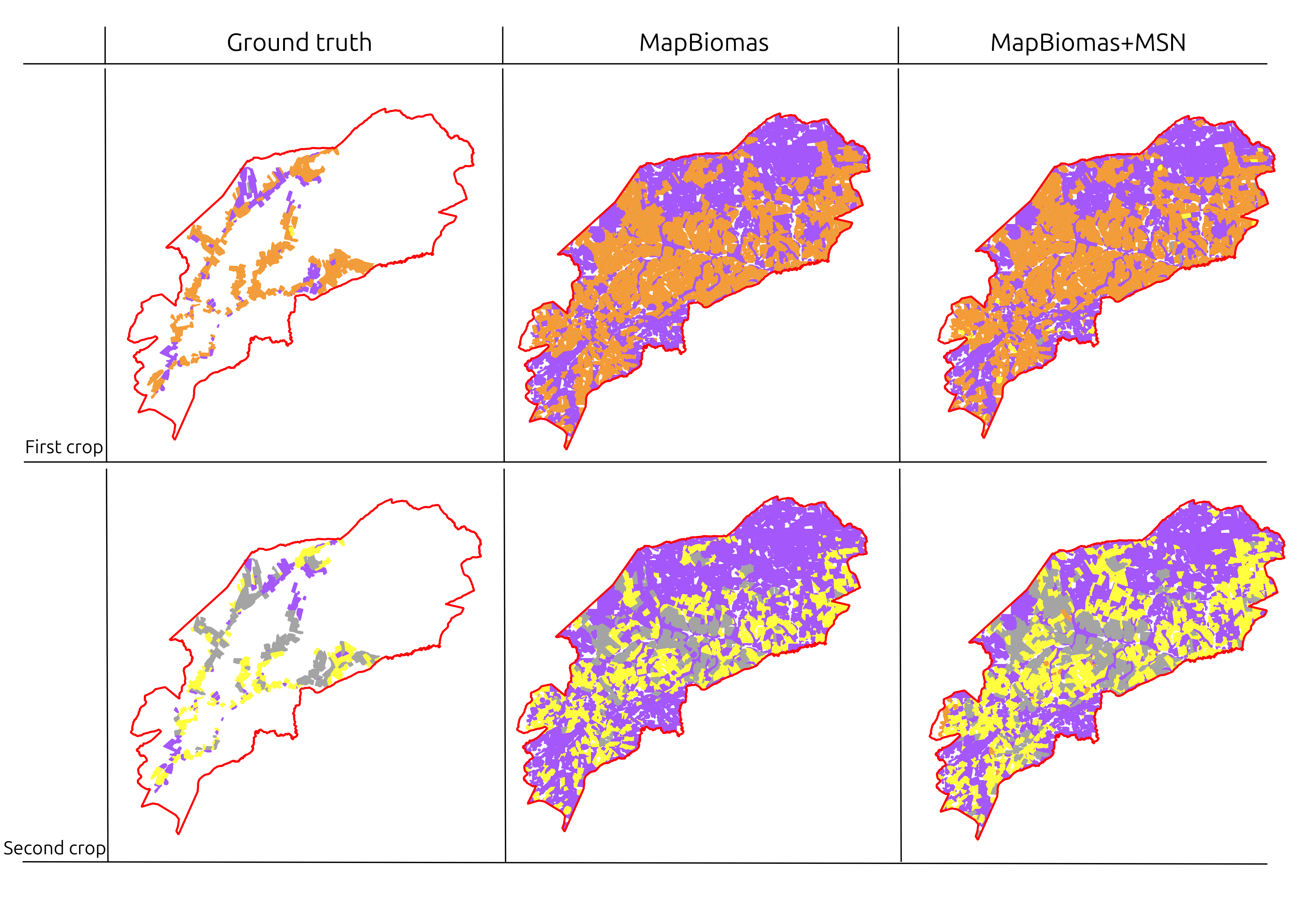}
    \caption{Crop field-level classifications for Campo Verde: reference, MapBiomas, and MapBiomas+MSN by crop season.}
    \label{fig:viz_CV_rotulado_masks_GERAL}
\end{figure}

\begin{figure}[H]
    \centering
    \includegraphics[draft=false, width=\linewidth]{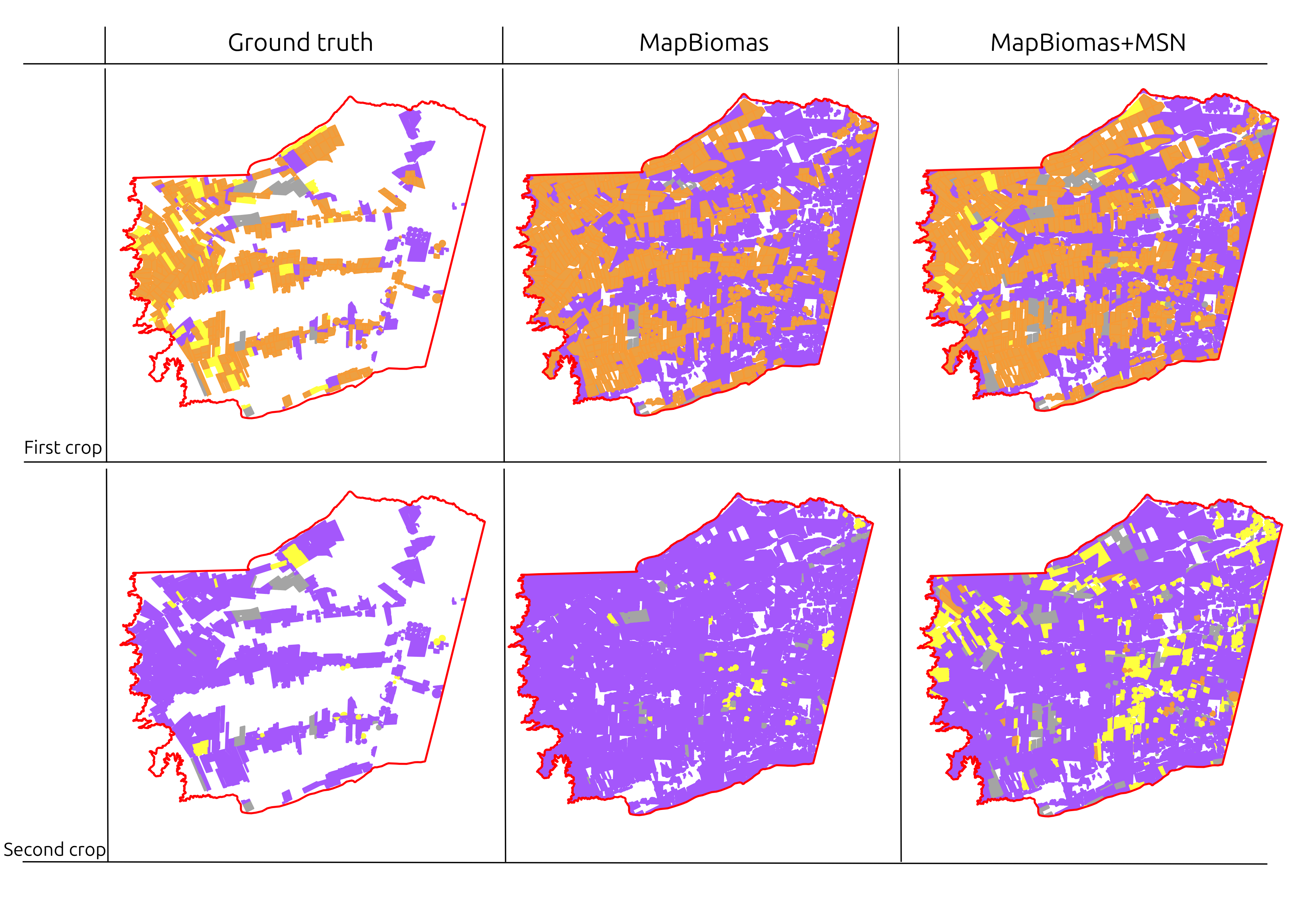}
    \caption{Crop field-level classifications for Luís Eduardo Magalhães: reference, MapBiomas, and MapBiomas+MSN by crop season.}
    \label{fig:viz_LEM_rotulado_masks_GERAL}
\end{figure}

In Luís Eduardo Magalhães (Table~\ref{tab:LEM_rotulado_mask}), the mean class accuracy gain reaches 9.42\% with 128 prototypes relative to MapBiomas, albeit with a modest reduction in overall accuracy. Figure~\ref{fig:LEM_matriz_area} reveals the origin of this asymmetry: MapBiomas correctly classifies only 40.3\% of the cotton area and 3.6\% of the maize area in LEM, while MSN 128 raises these values to 76.9\% and 9.6\%, respectively. The maize recall remains low in absolute terms, and the latent space analysis locates the reason. Maize is one of the best separated crops in the LEM embedding, with the second smallest intra-class dispersion in Figure~\ref{fig:distancia_msn} and a band of its own in Figure~\ref{fig:tsne_msn}, so the representation does isolate it. What is missing is the signal needed to name the corresponding clusters: with 64 maize patches in the entire region, MapBiomas rarely makes maize the predominant class of any cluster, and groups that the embedding separates end up named after another crop. This is a structural boundary of the mechanism rather than an incidental result, and it follows from its own design, since naming is delegated to the noisy source. When a class is nearly absent from that source, no amount of representation quality recovers it, and the gain observed for maize is bounded accordingly. Figures~\ref{fig:viz_CV_rotulado_masks_GERAL} and~\ref{fig:viz_LEM_rotulado_masks_GERAL} present the visual comparison of crop field-level classifications for both regions, showing that the proposed approach captures greater crop diversity relative to MapBiomas, especially in the second crop.

\begin{table}[h!]
\centering
\footnotesize
\caption{Crop field aggregation gain and comparison with the MapBiomas baseline: Campo Verde.}
\begin{tabular}{lcccccc}
\hline
\textbf{Model}
& \makecell{\textbf{OA} \\ \textbf{(Area-Based)}}
& \makecell{\textbf{MA} \\ \textbf{(Area-Based)}}
& \makecell{\textbf{$\Delta$ Gain} \\ \textbf{OA}}
& \makecell{\textbf{$\Delta$ Gain} \\ \textbf{MA}}
& \makecell{\textbf{$\Delta$ vs MB} \\ \textbf{OA}}
& \makecell{\textbf{$\Delta$ vs MB} \\ \textbf{MA}} \\
\hline
MSN - 16 prototypes   & 74.26\% & 76.63\% &  4.09\% &  4.06\% &  -7.03\% & -1.87\% \\
MSN - 32 prototypes   & 84.62\% & 82.34\% &  2.76\% &  2.84\% &   3.33\% &  3.84\% \\
MSN - 64 prototypes   & 85.63\% & 84.48\% &  2.64\% &  3.13\% &   4.34\% &  5.98\% \\
MSN - 128 prototypes  & 83.60\% & 82.00\% &  2.58\% &  2.96\% &   2.31\% &  3.50\% \\
MSN - 256 prototypes  & \textbf{87.74\%} & \textbf{86.08\%} & 3.42\% & 3.78\% & \textbf{6.45\%} & \textbf{7.58\%} \\
MSN - 512 prototypes  & 84.92\% & 83.57\% &  3.39\% & \textbf{4.04\%} &  3.63\% &  5.07\% \\
MSN - 1024 prototypes & 85.75\% & 84.34\% & \textbf{4.42\%} & \textbf{5.08\%} &  4.46\% &  5.84\% \\
MapBiomas     & 81.29\% & 78.50\% &  2.01\% &  2.33\% &      -   &    -    \\
\hline
\end{tabular}
\label{tab:CV_rotulado_mask}
\end{table}

\begin{table}[h!]
\centering
\footnotesize
\caption{Crop field aggregation gain and comparison with the MapBiomas baseline: Luís Eduardo Magalhães.}
\begin{tabular}{lcccccc}
\hline
\textbf{Model}
& \makecell{\textbf{OA} \\ \textbf{(Area-Based)}}
& \makecell{\textbf{MA} \\ \textbf{(Area-Based)}}
& \makecell{\textbf{$\Delta$ Gain} \\ \textbf{OA}}
& \makecell{\textbf{$\Delta$ Gain} \\ \textbf{MA}}
& \makecell{\textbf{$\Delta$ vs MB} \\ \textbf{OA}}
& \makecell{\textbf{$\Delta$ vs MB} \\ \textbf{MA}} \\
\hline
MSN - 16 prototypes   & 64.74\% & 60.13\% &  0.42\% &  0.41\% & -20.96\% &  1.97\% \\
MSN - 32 prototypes   & 79.20\% & 65.85\% &  1.75\% & \textbf{1.54\%} &  -6.50\% &  7.69\% \\
MSN - 64 prototypes   & 84.95\% & 67.56\% &  1.35\% &  0.75\% &  -0.75\% &  9.40\% \\
MSN - 128 prototypes  & 83.97\% & \textbf{67.58\%} & 1.05\% & 0.28\% & -1.73\% & \textbf{9.42\%} \\
MSN - 256 prototypes  & \textbf{85.27\%} & 66.23\% &  1.81\% &  0.87\% & \textbf{-0.43\%} &  8.07\% \\
MSN - 512 prototypes  & 85.22\% & 65.94\% &  1.53\% &  0.99\% &  -0.48\% &  7.78\% \\
MSN - 1024 prototypes & 84.70\% & 66.74\% & \textbf{1.90\%} &  1.48\% &  -1.00\% &  8.58\% \\
MapBiomas     & 85.70\% & 58.16\% &  0.93\% &  0.72\% &     -    &    -    \\
\hline
\end{tabular}
\label{tab:LEM_rotulado_mask}

\end{table}

Figures~\ref{fig:viz_CV_rotulado_masks_GERAL} and~\ref{fig:viz_LEM_rotulado_masks_GERAL} illustrate the scaling potential of the proposed methodology. The reference column highlights the sparsity of the annotated datasets, restricted to 628.72~km² in CV and 1,276.59~km² in LEM, while the MapBiomas and MapBiomas+MSN columns cover the entirety of each municipality: 3,758.08~km² and 3,344.19~km², respectively, representing expansions of approximately 6× and 2.6× without any additional annotation. Beyond the coverage gain, the visual comparison between the two prediction columns reveals qualitative improvement by the proposed approach, especially in the second crop, where minority crops such as cotton in CV gain greater representativeness in the classified fields, consistent with the substantial mean class accuracy gains documented in the previous tables. It should be noted that the magnitude of these improvements is sensitive to the quality of the noisy baseline: in LEM, where MapBiomas severely underestimates maize, the model starts from a weaker signal, and further gains will depend on improvements in the quality of the reference labels.

Before scaling the analysis to the state level, Table~\ref{tab:resumo_geral} consolidates the gain obtained over the labeled area, isolating the contribution of each mechanism. Starting from the MapBiomas baseline at the patch level, the \emph{semantic only} configuration applies the pseudo-group mechanism alone, the \emph{spatial only} configuration applies the crop field aggregation alone, and the \emph{combined} configuration applies both together; the semantic and combined rows adopt the best-performing configuration in each region (256 prototypes in CV and 128 in LEM). Combining both mechanisms yields the largest improvement in mean class accuracy, around 10 percentage points in both regions (from 76.17\% to 86.08\% in CV and from 57.44\% to 67.58\% in LEM), concentrated in the minority crops most underestimated by the baseline.

\begin{table}[h!]
\centering
\footnotesize
\caption{Effect of each mechanism over the labeled area: baseline, each mechanism alone, and both combined. OA: overall accuracy; MA: mean class accuracy.}
\label{tab:resumo_geral}
\begin{subtable}[t]{0.48\textwidth}
\centering
\caption{Campo Verde.}
\begin{tabular}{lcc}
\hline
\textbf{Configuration} & \textbf{OA} & \textbf{MA} \\
\hline
Baseline (MapBiomas) & 79.28\% & 76.17\% \\
Semantic only        & 84.32\% & 82.30\% \\
Spatial only         & 81.29\% & 78.50\% \\
\textbf{Combined}    & \textbf{87.74\%} & \textbf{86.08\%} \\
\hline
\end{tabular}
\label{tab:resumo_geral_cv}
\end{subtable}
\hfill
\begin{subtable}[t]{0.48\textwidth}
\centering
\caption{Luís Eduardo Magalhães.}
\begin{tabular}{lcc}
\hline
\textbf{Configuration} & \textbf{OA} & \textbf{MA} \\
\hline
Baseline (MapBiomas) & 84.77\% & 57.44\% \\
Semantic only        & 82.92\% & 67.30\% \\
Spatial only         & \textbf{85.70}\% & 58.16\% \\
\textbf{Combined}    & 83.97\% & \textbf{67.58\%} \\
\hline
\end{tabular}
\label{tab:resumo_geral_lem}
\end{subtable}
\end{table}

\begin{figure}[H]
    \centering
    \begin{subfigure}{0.48\textwidth}
        \centering
        \includegraphics[width=\linewidth]{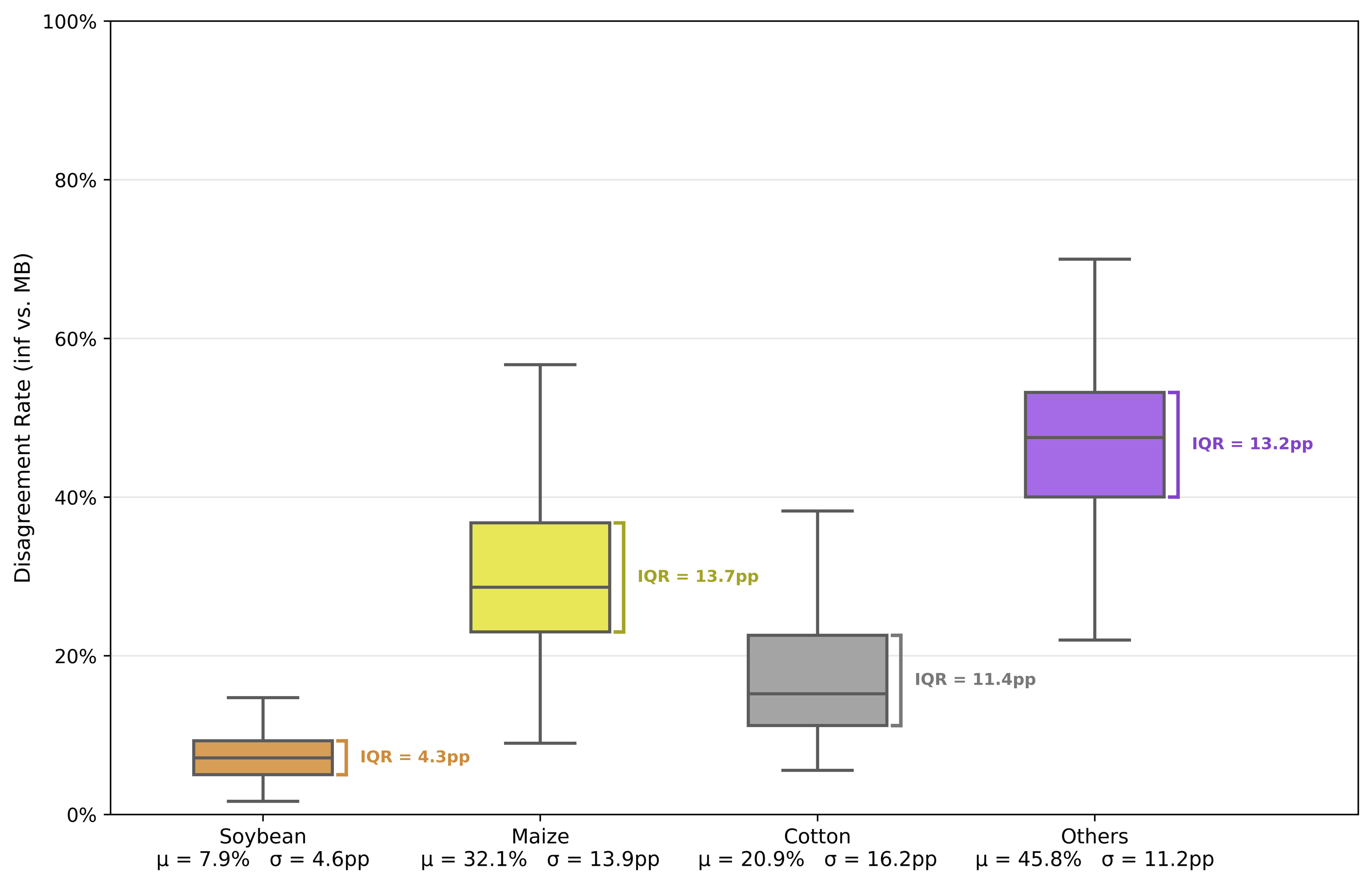}
        \caption{Mato Grosso.}
    \end{subfigure}
    \hfill
    \begin{subfigure}{0.48\textwidth} 
        \centering
        \includegraphics[width=\linewidth]{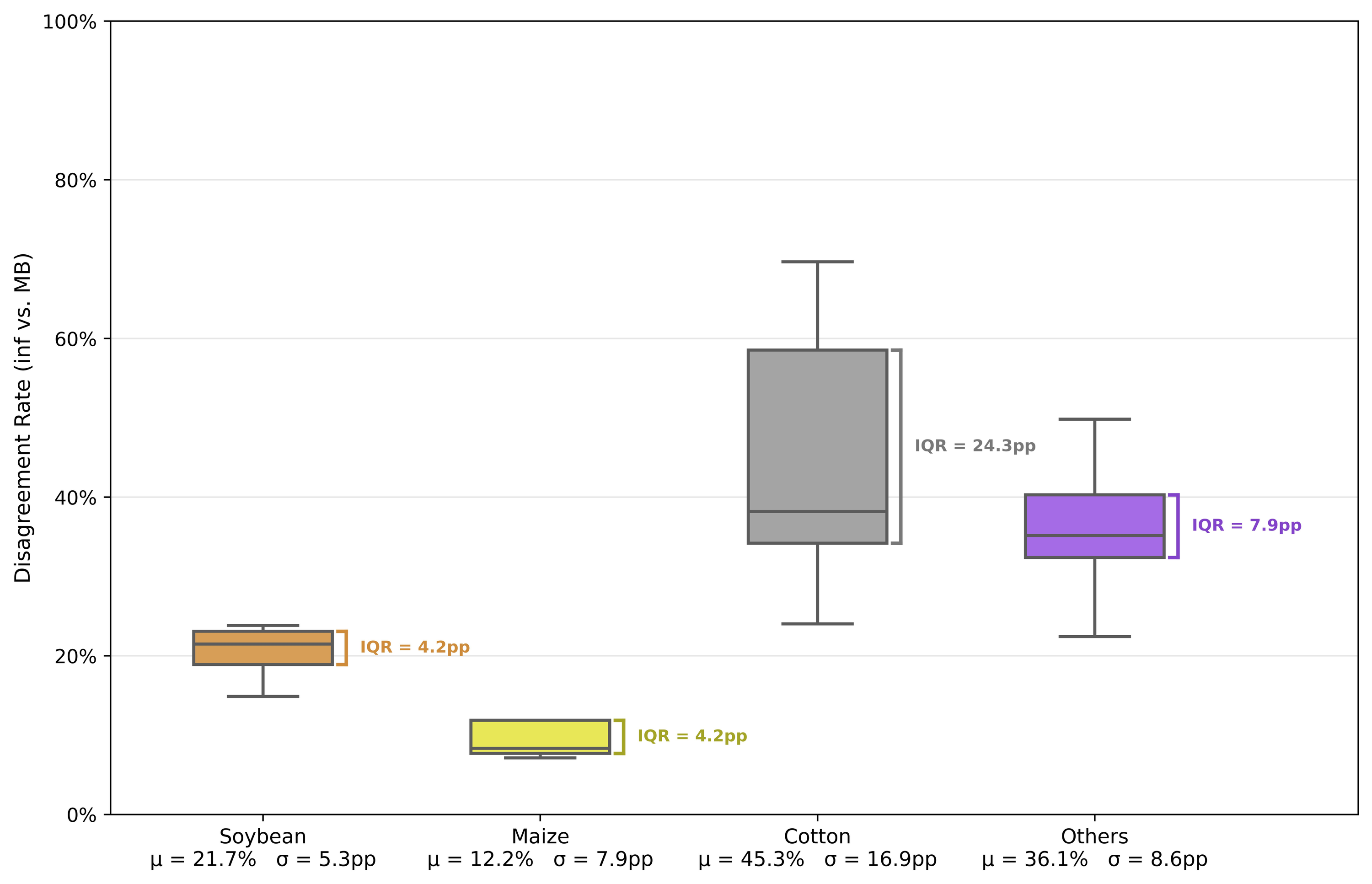}
        \caption{Bahia.}
    \end{subfigure}
    \caption{Per-class, per-municipality discordance between the proposed methodology and MapBiomas: Mato Grosso and Bahia (2020).}
    \label{fig:viz_estadual}
\end{figure}

To evaluate the state-scale generalization capacity of the methodology, Figure~\ref{fig:viz_estadual} presents the per-class, per-municipality discordance distribution for Mato Grosso and the agricultural region of Bahia. For soybean, a class well represented by MapBiomas, mean discordance is low and variance across municipalities is small, indicating stable behavior. For the remaining classes, absolute discordance is higher due to the known limitations of MapBiomas as a reference; however, the central point of the analysis is not the level of discordance but its spatial consistency. In both states, each class maintains a characteristic and stable discordance distribution across municipalities, with controlled variance even considering the heterogeneity of the evaluated regions. This behavior indicates that the proposed methodology generalizes in a predictable manner, and that further gains can be obtained as the quality of the reference labels improves.

\begin{figure}[H]
    \centering
    \includegraphics[draft=false, width=0.8\linewidth]{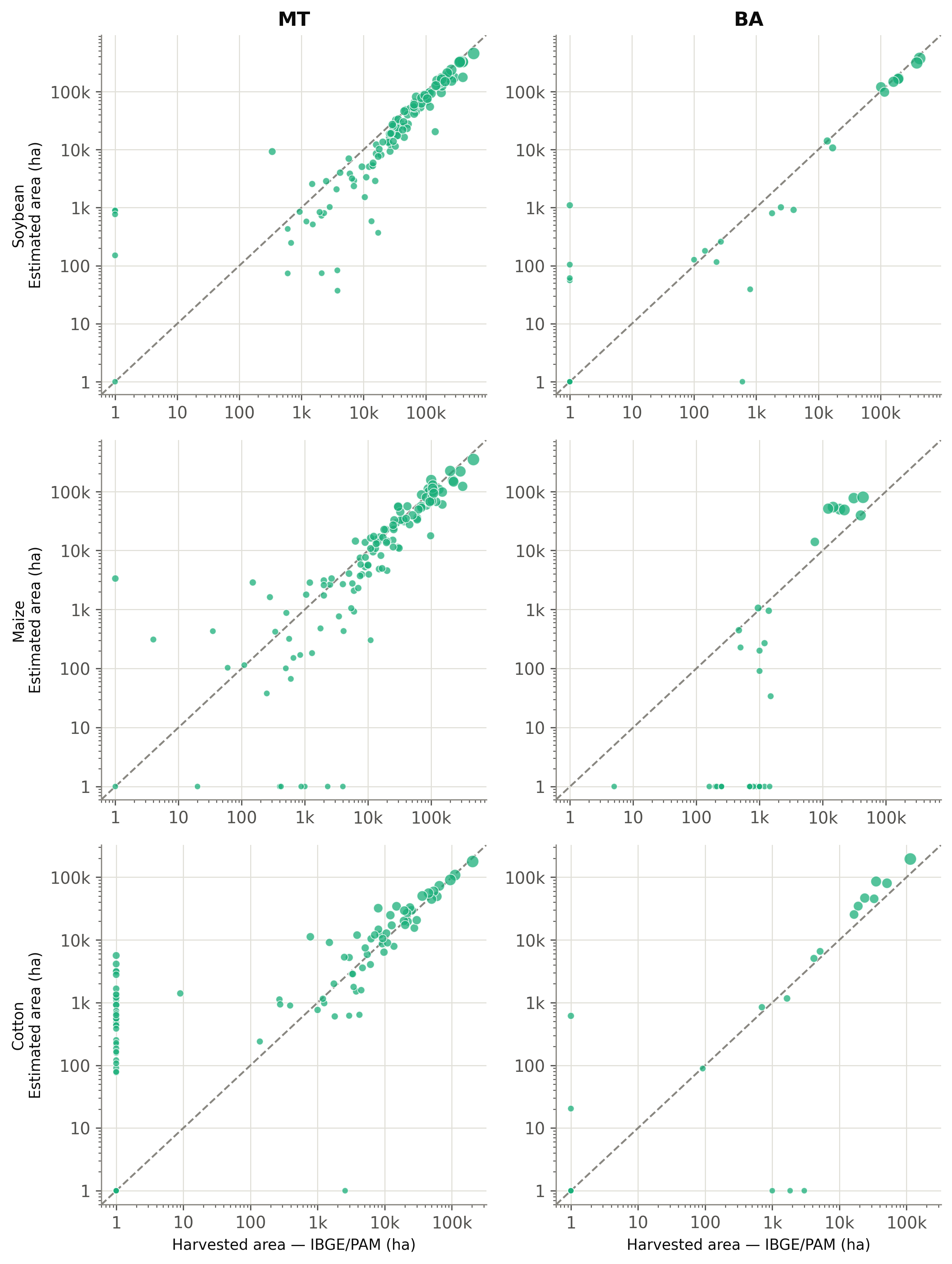}
    \caption{Municipal crop area estimated by the proposed methodology against the harvested area reported by IBGE/PAM for 2020, by crop (rows) and study area (columns).}
    \label{fig:concordancia_ibge}
\end{figure}

Every comparison reported so far confronts the pipeline with reference annotations or with MapBiomas, that is, with sources that are themselves derived from remote sensing. The last analysis of this section confronts it with an estimate of an entirely different nature. Figure~\ref{fig:concordancia_ibge} compares, municipality by municipality, the crop area estimated by the model in 2020 against the harvested area reported by IBGE/PAM for the same year, for soybean, maize, and cotton, in Mato Grosso (MT) and in Bahia (BA). Each point is a municipality, and both axes are logarithmic, since the areas involved range from a few hectares to hundreds of thousands. The size of the point is proportional to the number of segmented crop fields in the municipality, so that larger points correspond to estimates supported by more observations. The dashed diagonal is the 1:1 line, along which the two estimates coincide. Municipalities for which one of the two sources reports no area at all cannot be placed on a logarithmic scale and are pinned to a floor of one hectare; this is a display device and does not enter any computed value.

The two estimates are correlated in every panel: the correlation between them is 0.970 for soybean, 0.932 for maize, and 0.978 for cotton in Mato Grosso, and 0.997, 0.906, and 0.991 for the same crops in Bahia. More informative than the coefficient itself is where along the range the agreement is concentrated. In the municipalities in which a crop occupies large areas, in the upper right of each panel, the points collapse onto the 1:1 line and the two estimates become close. These are also the municipalities with the largest number of observed crop fields, and the ones in which the official estimate is itself most reliable, since the crop is economically prominent and locally well informed. Agreement is therefore strongest precisely in the regime in which either estimate would actually be used.

Dispersion grows as the area decreases, and reaches its extreme in the points pinned to the floor of the axes, where one of the two sources reports no area at all. This behavior is expected of both sources rather than of one of them: a few hundred hectares of a crop scattered over a large municipality are at once difficult to segment and classify from satellite imagery and difficult to estimate through a panel of local informants. The agreement thus degrades exactly where both estimates are individually weakest, which is the pattern one would expect of two independent descriptions of the same landscape, and not the pattern of a systematic failure on one side.

The purpose of this comparison is therefore not to evaluate accuracy: PAM is not ground truth, and the model is not a target for PAM. It is to verify that two estimates built from entirely disjoint evidence, one from satellite imagery and public land cover labels and the other from local declarations collected by official panels, describe the same agricultural landscape. They do, and they do so most closely where each crop is a substantial part of the municipality, which is the evidence sought here that the pipeline produces quantities of the right magnitude at the scale at which agricultural statistics are used.

\section{Conclusion}
\label{sec:conclusion}

This work proposed a methodology to build effective large-scale crop classification models directly from imperfect public land cover products, reducing simultaneously the semantic inconsistency of the labels and the spatial noise of the predictions. The two mechanisms act on these distinct sources of inconsistency: self-supervised pseudo-groups reduce the semantic inconsistency of the reference labels by shifting supervision from the sample level to the group level, while crop field-level aggregation reduces the spatial inconsistency of the predictions by exploiting the physical consistency of the agricultural field unit, suppressing intra-field noise and producing spatially coherent classifications. Evaluated independently and in combination in Campo Verde and Luís Eduardo Magalhães, both demonstrated consistent gains over MapBiomas and substantial recovery of minority crops systematically underestimated by the reference baseline. Consolidated over the labeled area (Table~\ref{tab:resumo_geral}), the two mechanisms combined raise mean class accuracy by 9.91 percentage points in Campo Verde and by 10.14 in Luís Eduardo Magalhães relative to the MapBiomas baseline, a gain that exceeds what either mechanism contributes in isolation and is concentrated in the minority crops most poorly represented by the reference.

A defining property of the methodology is that it relies exclusively on publicly available imagery and labels: it requires no new field campaign, no additional ground survey, and no manual label correction. This is what makes it inherently scalable. Automatic crop field segmentation by SAM2 enabled the expansion of the mapped area by up to six times without any annotation, and the state-scale generalization analysis, applied to entire states with no further model adjustment, revealed controlled discordance variance across dozens of municipalities, confirming that the methodology captures phenological patterns that transfer across regions. At that scale, the municipal crop areas derived from the predictions were also contrasted with the official Brazilian agricultural statistics (IBGE/PAM), an estimate produced from local declarations and therefore independent of every input used by the pipeline. The two are correlated for all three crops in both regions, and they converge onto the identity line precisely in the municipalities where a crop occupies a large area, which are also the ones in which each of the two estimates is individually most reliable. The official statistics are not ground truth, since they are estimated by panels of local informants rather than measured over the territory, so this is a consistency check and not an accuracy evaluation; that two estimates built from disjoint evidence describe the same agricultural landscape is nonetheless evidence that the pipeline produces quantities of the right magnitude at the aggregation level on which agricultural policy operates.

The analysis of the latent space shows that the self-supervised representation separates the crops well in both regions, with cluster purity reaching 89.0\% in Campo Verde and 91.2\% in Luís Eduardo Magalhães, independently of the noisy label source. The accuracy attained in each region is governed by the quality of the labels available to name the clusters rather than by the representation, so the magnitude of the gains is sensitive to the quality of the noisy reference baseline, and future improvements in MapBiomas versions are expected to amplify the results reported here. This same dependence explains the lower accuracy in Luís Eduardo Magalhães, where a crop such as maize is well isolated in the representation yet cannot be recovered by naming from the few examples the reference source provides.

A relevant limitation is the use of fixed temporal windows aligned to the regional average agricultural calendar. Intra-regional variations in planting timing may compromise the phenological representation of crop fields with early or late cycles. A direct extension of this work is the definition of adaptive windows at the crop field level, centered on the peak cultivation period of each parcel, which should particularly benefit the discrimination of second-crop species. The methodology is, by design, transferable to other remote sensing mapping domains where reliable labels are scarce.

\section*{Acknowledgment}

This study was funded by FAPERJ – Carlos Chagas Filho Foundation for the Support of Research in the State of Rio de Janeiro (Processes SEI 260003/013155/2024 and SEI 260003/006405/2024), by the National Council for Scientific and Technological Development (CNPq) (Brazil process number 420157/2023-9) and by the Getulio Vargas Foundation.


\bibliographystyle{elsarticle-harv}
\bibliography{references}

\end{document}